\documentclass[a4paper,10pt]{article}
\usepackage{bbding}
\usepackage{subcaption}
\usepackage{graphicx} 
\usepackage{natbib}
\usepackage{hyperref}
\usepackage{settings}
\usepackage{tikz-cd}
\usetikzlibrary{3d, arrows.meta, decorations.pathmorphing}

\title{What is a good matching of probability measures? A counterfactual lens on transport maps}
\author{
  Lucas De Lara\thanks{Institut Elie Cartan de Lorraine, Université de Lorraine, Lorraine, Grand Est region, France.}
  \and
  Luca Ganassali\thanks{Université Paris-Saclay, CNRS, Inria, Laboratoire de mathématiques d'Orsay, Orsay, Essonne, Île-de-France region, France.}
}

\date{\today}

\begin{document}

\maketitle

\begin{abstract}
Coupling probability measures lies at the core of many problems in statistics and machine learning, from domain adaptation to transfer learning and causal inference. Yet, even when restricted to deterministic transports, such couplings are not identifiable: two atomless marginals admit infinitely many transport maps. The common recourse to optimal transport, motivated by cost minimization and cyclical monotonicity, obscures the fact that several distinct notions of multivariate monotone matchings coexist.

In this work, we first carry a comparative analysis of three constructions of transport maps: cyclically monotone, quantile-preserving and triangular monotone maps. We establish necessary and sufficient conditions for their equivalence, thereby clarifying their respective structural properties. 

In parallel, we formulate counterfactual reasoning within the framework of structural causal models as a problem of selecting transport maps between fixed marginals, which makes explicit the role of untestable assumptions in counterfactual reasoning.

Then, we are able to connect these two perspectives by identifying conditions on causal graphs and structural equations under which counterfactual maps coincide with classical statistical transports. In this way, we delineate the circumstances in which causal assumptions support the use of a specific structure of transport map.
Taken together, our results aim to enrich the theoretical understanding of families of transport maps and to clarify their possible causal interpretations. We hope this work contributes to establishing new bridges between statistical transport and causal inference.
\end{abstract}


\section{Introduction} 

\paragraph{Coupling probability measures}
Many statistical applications such as domain adaptation \citep{courty2016optimal,redko2017theoretical,xu2020reliable,chang2022unified} and transfer learning \citep{gayraud2017optimal,peterson2021transfer} require coupling two probability measures. This fundamental problem faces an indeterminacy issue---even when focusing on deterministic couplings, namely, couplings that concentrate on the graphs of transport maps. As a matter of fact, two atomless marginal probability distributions admit infinitely many transport maps. This raises the question of picking an adequate function among all these choices. 
Optimal transport \citep{monge1781memoire} has become an established framework to define matchings of measures, due to its strong theoretical and practical developments \citep{villani2008optimal,peyré2020computationaloptimaltransport}. The following quote from \cite{manole2024plugin} upholds this idea:
\begin{displayquote}
\say{A wide range of statistical applications involve transforming random variables to ensure they follow a desired distribution. Optimal transport maps form natural choices of such transformations when no other canonical choice is available.}
\end{displayquote}
The recognition of optimal transport maps for the squared Euclidean cost comes from two valued features: they minimize a standard transport cost and they are cyclically monotone \citep{cuesta1989notes,brenier1991polar}. Interestingly, cyclical monotonicity generalizes the notion of non-decreasing function to higher dimensions, which led several scholars to argue that optimal transport was the relevant approach to extend the classical univariate monotone matching to the multivariate case \citep{charpentier2023optimal,torous2024optimal,delara2024transport}.

In our work, we notably challenge the idea the optimal transport maps are the most suited options to do so. While the monotone rearrangement between one-dimensional marginals satisfies an intuitive rank-preserving property, it does not precisely generalize to higher dimensions, due to the absence of canonical ordering. 

In particular, there exist various possible constructions of \say{rank-preserving} transports that collapse in dimension one but differ in general: we briefly highlight three possibilities. First, one could rely on the unique cyclically monotone transport map \citep{mccann1995}, which matches globally the two marginals. Second, one could take the unique triangular monotone increasing transport map \citep{rosenblatt1952remarks}, which successively matches unidimensional disintegrations of the the marginals, introducing a hierarchy over the variables. Third, one could choose a notion of multivariate quantiles \citep{hallin2021distribution}, and then match the marginals via multivariate quantile preservation. This coexistence motivates further discussions on multivariate monotonic transport maps and further studies on alternatives to optimal transport when it comes to choosing a good matching of measures.  

\paragraph{Link with counterfactual couplings}
One of the original motivations for this work comes from causal inference, where a recent line of work suggests to define counterfactual couplings directly via transport maps \citep{black2020fliptest,charpentier2023optimal,torous2024optimal,delara2024transport,machado2025sequential,balakrishnan2025conservative}, circumventing the specification of a causal model. Let us briefly recall the context and explain the terminology. In causal inference, one can determine via randomized experiments whether a medical treatment works in general, but cannot determine whether Alice specifically would have been cured had her treatment assignment changed. As such, reasoning at the counterfactual level (the sharpest layer of causation, which concerns contrary-to-fact statements) requires making untestable choices of matchings between distributions of outcomes from several parallel worlds \citep{lewis2013counterfactuals,pearl2018book}. 
The unidentifiability of couplings from their marginals discussed before also arises in this context and renders general counterfactual reasoning unfalsifiable without additional assumptions. While this unfalsifiability led \cite{dawid2000causal} to consider counterfactual reasoning as unscientific, counterfactuals are useful to notably define fundamental concepts of fairness \citep{kusner2017counterfactual}, harm \citep{richens2022counterfactual,mueller2023personalized,sarvet2025perspectives}, and credit \citep{mesnard2021counterfactual}. Thereby, clarifying the choices that underpin the evaluation of counterfactual statements represents a fundamental problem with practical applications.

In this article, we propose to relate the statistical and counterfactual perspectives on transport maps. In particular, we discuss the relation between a certain choice of transport map interpreted as a counterfactual matching and the latent causal model formalizing the causal assumptions. More specifically, we rely on the framework of \cite{pearl2009causality}, where causal assumptions are framed in terms of causal graphs and structural equations. 

\paragraph{Paper outline}
The rest of the paper proceeds as follows. In \cref{sec:transport}, we provide details on the three aforementioned ways of matching probability measures from a purely statistical angle. Our comparative analysis enables us to establish sufficient and necessary conditions for these transport to coincide. We point out that our study does not focus on numerical aspects. In \cref{sec:counterfactual}, we thoroughly specify a framework for counterfactual reasoning, based on structural causal models, in which counterfactuals emerge from transport maps between fixed marginals. \cref{sec:meet} bridges the preceding two sections by employing the unified transport-map formalism to investigate potential connections between the classical transport maps introduced in \cref{sec:transport} and the counterfactual maps discussed in \cref{sec:counterfactual}. Specifically, we establish necessary and/or sufficient conditions on the structural causal model under which the counterfactual maps can be characterized as optimal transport maps, quantile-preserving maps, or Knothe--Rosenblatt transports. We provide precise theoretical results that delineate when such equivalences hold, and discuss the employment of the classical transport maps in causal tasks. Finally, \cref{sec:further} is dedicated to further discussions.\\

In doing this work, we aim at enriching the explanatory basis of transport maps via analytical properties as well as causal interpretations. 
We hope that this article will serve as a bridge between two communities: on the one hand, it can guide statisticians who employ transport methods to engage with causal reasoning and to recognize which constructions are appropriate—or problematic—in that framework; on the other hand, it can help researchers in causal inference to elucidate and articulate more clearly the connections between counterfactuals and transport.

\subsection*{Basic notation}
Throughout the article, let $(\Omega,\Sigma,\bbP)$ be an abstract probability space on which all random variables will be defined. A random variable (including random vectors) is a measurable function from $\Omega$ to an Euclidean space equipped with the Borel $\sigma-$algebra. For $d \geq 1$ an integer, $\calP(\R^d)$ refers to the set of Borel probability measures on $\R^d$. The law of any random variable $X : \Omega \to \R^d$ is denoted by $\law{X} \in \calP(\R^d)$. For some $P \in \calP(\R^d)$, we also use the notation $X \sim P$ to mean that $\law{X} = P$. 

For any Borel measure $\mu$ on $\R^d$ (not necessarily a probability measure) and any Borel set $E$ of $\R^d$, we say that $E$ holds $\mu-$almost everywhere ($\mu-$a.e.) if $\mu(\R^d \setminus E)=0$. For two functions $f,g$ on $\R^d$, we write $f=g$ $\mu-$a.e. if $\mu(\left\{ x \in \R^d \mid f(x) \neq g(x)\right\})=0$. 

Moreover, we say that $P \in \calP(\R^d)$ has a density if it is absolutely continuous with respect to the Lebesgue measure. In particular, if $P$ has a density then it is atomless. The relation $X \independent Y$ signifies that the random variables $X$ and $Y$ are independent. The symbol $\otimes$ indicates the product of probability measures. We write $\law{Y | X=x}$ for the conditional law of $Y$ given $X=x$, which is well-defined for $\law{X}-$almost every $x$.  

For $\mu \in \calP(\R^d)$ and $T: \R^d \to \R^d$ a Borel function, the pushforward measure of $\mu$ by $T$, denoted $T \sharp \mu$, is the measure of $\calP(\R^d)$ defined as follows: 
    $$ \mbox{for all Borel set $B$, } T \sharp \mu(B) = \mu(T^{-1}(B)) \, . $$

Let $(\X_i)_{i \in I}$ be a collection of spaces indexed by a finite set $I$. For any subset $J \subseteq I$, we write $\X_J$ for the Cartesian product $\times_{i \in I \cap J} \X_i$. By convention, $\X_J = \varnothing$ if $I \cap J = \varnothing$. Similarly, for any tuple $(x_i)_{i \in I} \in \X_I$, we define $x_J = (x_i)_{i \in I \cap J} \in \X_J$. Given a family of functions $(f_i)_{i \in I}$ such that $f_i : \X \to \Y_i$, we define $f_I : \X \to \Y_I, x \mapsto (f_i(x))_{i \in I}$. Given a family of functions $(f_i)_{i \in I}$ such that $f_i : \X_i \to \Y_i$, we define $\times_{i \in I} f_i : \X_I \to \Y_I, x \mapsto (f_i(x_i))_{i \in I}$. The cardinality of a set $I$ is denoted by $\abs{I}$.

With a slightly abusing notation, we denote by $\Id$ the identity function and the identity matrix, whatever the considered space. Moreover, we write $\langle \cdot , \cdot \rangle$ and $\norm{ \cdot }$ for, respectively, the Euclidean scalar product and the Euclidean norm, regardless of the dimension. For every matrix $M \in \R^{d \times d}$, every vector $b \in \R^d$, and any $I,J \subseteq [d]$, we define $M_{I,J} := (M_{i,j})_{i \in I,j \in J} \in \R^{\abs{I} \times \abs{I}}$ and $b_I := (b_i)_{i \in I} \in \R^{\abs{I}}$.

The \emph{gradient} and the \emph{Hessian} of a twice differentiable function $\varphi : \R^d \to \R$ are denoted by, respectively, $\nabla \varphi : \R^d \to \R^d$ and $\nabla^2 \varphi : \R^d \to \R^{d \times d}$. We recall that $\nabla^2 \varphi(x)$ is symmetric for every $x \in \R^d$. The \emph{Jacobian} of a differentiable function $f : \R^d \to \R^d$ is denoted by $\Jac(f): \R^d \to \R^{d \times d}$.

\section{Three ways of matching probability measures}\label{sec:transport}

In this section, after recalling generalities on transport maps, we introduce three different canonical transport maps, each of which will be motivated by applications from various fields, which are -- purposely -- not related to causality or counterfactual reasoning. We emphasize that our comparison essentially focuses on theoretical aspects, although we will briefly mention computational ones.

\subsection{Generalities}

Let $\mu, \nu \in \calP(\R^d)$ and $X \sim \mu$. The general question which motivates the definition of transport maps is the following: \emph{how can one transform the random variable $X$ to ensure that it follows the target distribution $\nu$?}

Let $\mu, \nu \in \calP(\R^d)$. A Borel function $T: \R^d \to \R^d$ is a \emph{transport map from $\mu$ to $\nu$} if $T \sharp \mu = \nu$. We denote by

$$ \T(\mu,\nu) :=  \{ T: \R^d \to \R^d, \mbox{ Borel}, \quad T \sharp \mu = \nu\}, $$
the set of transport maps from $\mu$ to $\nu$. In terms of random vectors, if $T \in \T(\mu,\nu)$ then $T(X) \sim \nu$ whenever $X \sim \mu$.

In the usual formulation, $\mu$ and $\nu$ are either known or can be approximated -- e.g. via i.i.d. samples from $\mu$ and $\nu$ at our disposal -- and we are interested in finding a transport map. However, the objective can be flipped in some applications, where a transport map previously computed can be used to obtain samples from a distribution that was otherwise out of reach for the statistician -- see the use of transport maps in Nonlinear Difference-in-Differences \cite{athey_CIC_06, torous2024optimal}.

\begin{remark}\label{rk:mu_ae_only}
    Note that for $T$ to be a transport map from $\mu$ to $\nu$, it is enough to define $T$ $\mu-$a.e, rather than over the whole space $\R^d$. In the sequel we may encounter such maps which are only defined $\mu-$a.e., but still call them transport maps.
\end{remark}

Beforehand, let us state a basic lemma on push-forward measures.

\begin{lemma}[Basic push-forward calculus]\label{lem:pushforward}
Let $\mu, \nu \in \calP(\R^d)$, and $T_1,T_2 : \R^d \to \R^d$ be Borel functions. Then,
\begin{enumerate}
    \item[$(i)$] $(T_1 \circ T_2) \sharp \mu = T_1 \sharp (T_2 \sharp \mu)$;
    \item[$(ii)$] if $T_1$ is bijective, then $T_1^{-1}$ is also Borel and $\nu = T_1 \sharp \mu \iff \mu = T_1^{-1} \sharp \nu$.
\end{enumerate}
\end{lemma} 

Note that for given distrubutions $\mu, \nu \in \calP(\R^d)$, $\T(\mu,\nu)$ may be empty (take $\mu = \delta_x$ and $\nu = \frac{1}{2}\delta_x + \frac{1}{2}\delta_y$ with $x \neq y \in \R^d$), or infinite (for example, any orthogonal operator in $\R^d$ is a transport map from the isotropic Gaussian $\calN(0,\Id)$ to itself).

Next, we will focus on marginals $\mu, \nu$ have densities. In this configuration, the set of transport maps is always infinite uncountable (as a special case of \citep[Proposition 3.4]{delara2025nonconvexity}). As mentioned in the introduction, this raises the question of which transport map is the more appropriate.

In the following, we provide three different criteria that one may want a transport map to satisfy. We highlight their motivations depending on the context, their properties, and provide their corresponding transport maps.

\subsection{Cyclically monotone transport map}

A first property that one may want to impose to the transport map is to be non-decreasing. To illustrate this in dimension one, think of a physical system where $\mu$ and $\nu$ represent mass distributions of a given set of particles along a line at two different instants. When considering displacing the particles from $\mu$ to $\nu$, it is physically inefficient and unnatural for them to leapfrog over each other. A non-decreasing map ensures that particles move without crossing paths, which minimizes interference and energy cost.

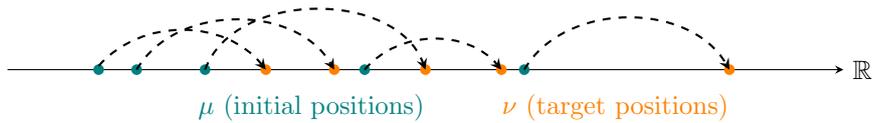
\begin{figure}[h]
    \centering
    \begin{tikzpicture}[>=stealth]
  \draw[->] (-0.5, 0) -- (10.5, 0) node[right] {$\mathbb{R}$};
  \fill[teal] (0.7, 0) circle (2pt);
  \fill[teal] (1.2, 0) circle (2pt);
  \fill[teal] (2.1, 0) circle (2pt);
  \fill[orange] (2.9, 0) circle (2pt);
  \fill[orange] (3.8, 0) circle (2pt);
  \fill[teal] (4.2, 0) circle (2pt);
  \fill[orange] (5.0, 0) circle (2pt);
  \fill[orange] (6.0, 0) circle (2pt);
  \fill[teal] (6.3, 0) circle (2pt);
  \fill[orange] (9.0, 0) circle (2pt);
  \draw[->, dashed, bend left=50, thick] (0.7, 0.05) to (2.9, 0);
  \draw[->, dashed, bend left=60, thick] (1.2, 0) to (3.8, 0);
  \draw[->, dashed, bend left=70, thick] (2.1, 0) to (5.0, 0);
  \draw[->, dashed, bend left=50, thick] (4.2, 0) to (6.0, 0);
  \draw[->, dashed, bend left=60, thick] (6.3, 0) to (9.0, 0);
  \node[teal] at (3.5, -0.5) {$\mu$ (initial positions)};
  \node[orange] at (7.5, -0.5) {$\nu$ (target positions)};
\end{tikzpicture}
    \caption{A non-decreasing transport map in 1D.}
    \label{fig:CM_1D}
\end{figure}

When $d>1$, a natural notion which extends non-decreasing maps from $\R^d$ to $\R^d$ is given by cyclic monotonicity. 

\begin{definition}[Cyclic monotonicity]
    A map $T: \R^d \to \R^d$ is \emph{cyclically monotone (CM)} if for any $m \geq 1$ and any $x^{(1)}, x^{(2)}, \ldots, x^{(m)} \in \R^d$,
    $$ \sum_{i=1}^{m} \langle x^{(i)}-x^{(i+1)}, T(x^{(i)}) \rangle = \sum_{i=1}^{m} \langle x^{(i)}, T(x^{(i)})-T(x^{(i+1)}) \rangle \geq 0,$$ where $x^{(m+1)} = x^{(1)}$. 
\end{definition}
We refer to \cite{depascale202460yearscyclicmonotonicity} for an exhaustive survey on cyclic monotonicity. Note that as claimed, this definition indeed meets that of non-decreasing functions when $d=1$. 

\begin{lemma}\label{lem:CM_is_M_in_1D}
A map $T: \R \to \R$ is cyclically monotone if and only if $T$ is non-decreasing.
\end{lemma}

Besides, note that if $\varphi : \R^d \to \R$ is a differentiable convex function, then its gradient $\nabla \varphi: \R^d \to \R^d$ is cyclically monotone. Indeed, by convexity, for all $x^{(i)},x^{(i+1)} \in \R^d$, $\varphi(x^{(i+1)}) - \varphi(x^{(i)}) \geq \langle x^{(i+1)}-x^{(i)}, \nabla \varphi(x^{(i)}) \rangle$ and summing over $1 \leq i \leq m$ gives the desired result. Rockafellar’s theorem establishes the converse result (Theorem 24.8 in \cite{rockafellar1997convex}), showing that $T : \R^d \to \R^d$ is cyclically monotone if and only if $T$ coincides Lebesgue-almost everywhere on $\R^d$ with the gradient of a closed, proper convex function $\varphi: \R^d \to \R \cup \{ +\infty\}$\footnote{To be more precise, the converse result is stated and holds for a more general definition of cyclic monotonicity, allowing for sets instead of maps. Theorem 24.8 in \cite{rockafellar1997convex} states that $\Gamma \subset \R^d \times \R^d$ is CM if and only if $\Gamma$ is included in the subdifferential $\partial \varphi$ of a convex function $\varphi: \R^d \to \R \cup \{ + \infty\}$. In our case, since $\Gamma$ is of the form $\{ (x,T(x)), x \in \dE \}$, $\partial \varphi(x)$ has to be non empty for all $x \in \dE$, and thus $\varphi < \infty$ on $U$. $\varphi$ is finite, convex on $U$: the set of its non-differentiability points is of Hausdorff dimension at most $d-1$. Thus, $T$ coincides with $\nabla \varphi$ Lebesgue-a.e. on $\dE$. }

A key theorem due to \cite{mccann1995} establishes that if the source measure $\mu$ has a density, a CM transport map in exists and is unique up to changing it on $\mu-$null measure sets. 

\begin{theorem}[\cite{mccann1995}]\label{thm:mccann}
    Let $\mu,\nu \in \calP(\R^d)$. Assume that $\mu$ has a density. Then, their exists a closed, proper convex function $\varphi: \R^d \to \R \cup \{ + \infty \}$ which gradient $\nabla \varphi$ is such that $(\nabla \varphi)\sharp \mu=\nu$. Moreover, this map is the unique cyclically monotone transport map in $\T(\mu,\nu)$, up to changing it on $\mu-$null measure sets. 
\end{theorem}

Note that in the above theorem, since $\varphi$ is convex and $\mu$ has a density, $\varphi$ is differentiable everywhere but on a $\mu-$null measure set and thus $\nabla \varphi$ is well defined as a transport in view of \cref{rk:mu_ae_only}. By convention, in the context of \cref{thm:mccann}, such a map $\nabla \varphi$, on its domain, will be called \emph{the cyclically monotone transport map} from $\mu$ to $\nu$, and denoted 
$$\CM(\mu, \nu) := \nabla \varphi \, .$$
All other CM transport maps in $\T(\mu,\nu)$ will be obtained from $\CM(\mu, \nu)$, changing it arbitrarily on $\mu-$null measure sets.

Now that $\CM(\mu, \nu)$ is defined properly, let us study its invertibility. Unsurprinsingly, when $\CM(\mu, \nu)$ and $\CM(\nu, \mu)$ exist, they are a.e. inverses of each other. The following proposition is standard but the proof is given for the sake of completeness, in a compact way.

\begin{proposition}[Inversion of CM maps]\label{prop:OT_invertibility}
    Assume that $\mu, \nu \in \calP(\R^d)$ both have densities. Then,  
    $$ \mu\mbox{-a.e.}, \; \CM(\nu, \mu) \circ \CM(\mu, \nu) = \Id \quad \mbox{and} \quad \nu\mbox{-a.e.}, \; \CM(\mu, \nu) \circ \CM(\nu, \mu) = \Id.$$
\end{proposition}

\paragraph{CM maps and optimal transport.}
A striking fact is that imposing cyclic monotonicity on the transport map not only captures inherent constraints present in many physical and economic systems, but also results, under additional assumptions, in an optimal transport map for the $L^2$ cost. Celebrated results (see \citep{cuesta1989notes,brenier1991polar}) establish that under the assumptions of \cref{thm:mccann}, if moreover $\mu, \nu$ admit finite second moments, then the CM transport map $\CM(\mu, \nu)$ is also the $\mu-$a.e. unique solution to the \emph{optimal transport problem}:
    \begin{equation}\label{eq:OT}
     \OT(\mu,\nu) = \argmin_{T \in \T(\mu,\nu)} \int \norm{x-T(x)}^2 \mathrm{d} \mu(x) \, .
     \end{equation} 
In this case, $\CM(\mu,\nu)$ is frequently referred to as the \emph{Brenier map} from $\mu$ to $\nu$. We emphasize that the existence of the CM transport map does \emph{not} relate to optimal transport. We refer to \cite{villani2008optimal,peyré2020computationaloptimaltransport} for a general overview of optimal transport. 

The above interpretation of $\CM(\mu, \nu)$ as $\OT(\mu,\nu)$ not only sheds light on the properties of a CM transport (it satisfies an optimality criterion) but also provides practical algorithms to estimate it from data. A common strategy consists of employing solvers of \cref{eq:OT} between empirical versions of $\mu$ and $\nu$ (see \citep{flamary2021pot}), and then to extends the discrete solution on the whole domain through regularity assumptions \citep{hallin2021distribution,delara2021consistent,manole2024plugin,beirlant2020center}. Another popular approach parametrizes the transport map (or the convex potential) as a neural network, to efficiently generalize to the matching to new samples \citep{makkuva2020optimal,korotin2021neural,korotin2021wasserstein,huang2021convex,gonzalez2022gan,korotin2023neural}.

\subsection{Quantile-preserving transport map}
A second property that one may want the transport map to meet is quantile preservation. The motivation arises when considering ranked decision-making scenarios. To illustrate this in dimension one, suppose that we want to select students for a PhD program based on their aggregated GPA. Imagine one has to decide between two students, one coming from institution $A$, the other coming from an equivalent institution $B$. Since students come from different institutions, the raw scores may differ in scale or spread across students. So how to compare them? 

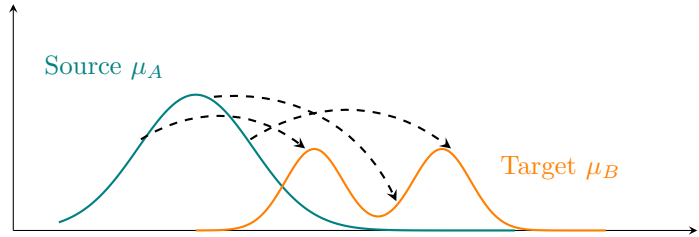
\begin{figure}[h]
    \centering
    \begin{tikzpicture}[>=stealth, scale=1.2]
        \draw[->] (0, 0) -- (7.5, 0) node[right] {};
        \draw[->] (0, 0) -- (0, 2.5) node[above] {};
        \draw[teal, thick, domain=0.5:5.5, samples=100, smooth] 
            plot (\x, {1.5*exp(-((\x - 2.0)^2)/0.8)}) node[teal, right] {};
        \draw[orange, thick, domain=2:6.5, samples=200, smooth] 
            plot (\x, {
                0.9*exp(-((\x - 3.3)^2)/0.2) + 
                0.9*exp(-((\x - 4.7)^2)/0.2)
            }) node[orange, right] {};
        \foreach \x/\y in {1.4/3.2, 2.2/4.2, 2.6/4.8} {
            \draw[->, thick, dashed, bend left=35] 
                (\x, {1.5*exp(-((\x - 2.0)^2)/0.8)+0.05}) 
                to 
                (\y, {0.9*exp(-((\y - 3.3)^2)/0.2) + 0.9*exp(-((\y - 4.7)^2)/0.2)+0.05});
        }
        \node[teal] at (1.0, 1.8) {Source $\mu_A$};
        \node[orange] at (6, 0.7) {Target $\mu_B$};
    \end{tikzpicture}
    \caption{Quantile preserving transport from $\mu_A$ (GPAs in institution A) to $\mu_B$ (GPAs in institution B).}
    \label{fig:bimodal_transport}
\end{figure}

If the differences between institutions are due to their environments rather than the students themselves, what really matters in the selection process is each candidate’s relative position within their own institution. For example, we want to ensure that two students who rank in the top 5\% at their respective institutions have an equal opportunity.

This is exactly the property of the quantile-preserving map: it recalibrates the grade of the student from institution $A$ based on their rank within its group, mapping it to the equivalent grade of a student with same rank in institution $B$. This process makes the selection process arguably fair and interpretable, allowing for consistent comparison across varying distributions.\\

When $d=1$, the quantile associated to a level $\alpha \in ]0,1[$ of a distribution with c.d.f. $F$, is given by 
\begin{equation}\label{eq:quantile_1D}
    F^{\dagger}(\alpha) := \inf \{ x \in \R, \, \alpha \leq F(x)\} \, .
\end{equation} The quantile function is always non-decreasing.
If $F$ is invertible (that is, strictly increasing on some domain) then $F^{\dagger}$ and $F^{-1}$ coincide. For two distributions $\mu$ and $\nu$, denoting by $F_\mu$ and $F_\nu$ their c.d.f.s, the usual quantile-preserving transport map from $\mu$ to $\nu$ is given by $F^{\dagger}_{\nu} \circ F_{\mu} $. 

When $d \geq 2$, multivariate quantiles have been the focus of a vast research literature. We refer the reader to \cite{serfling02, hallin2021distribution} as well as the introduction of \cite{thurin24these} for an general overview.

A tractable and interpretable definition of multi-dimensional quantiles is proposed in \cite{chernozhukov2015}: the \emph{Monge-Kantorovich quantiles}. These are obtained by transporting a reference measure $p_0 \in \calP(\R^d)$ to the distribution of interest via a CM map: if $\CM(\mu,p_0)(x) = \alpha \in \R^d$, then $x$ is the $p_0-$quantile of order $\alpha$ for $\mu$.

As a special case, \cite{hallin2021distribution} propose the \emph{center-outward} definition of quantile functions, which are the Monge-Kantorovich quantiles for the reference measure $p_0 = \Circle$, where $\Circle$ is the spherical uniform distribution\footnote{Defined by the law of $R \cdot U$, with $R \sim \Unif([0,1])$ and $U \sim \Unif(S_{d-1})$ independent, where $S_{d-1}$ is the unit sphere of $\R^d$.} over the $d-$dimensional unit ball $B_d$. If $x \in \R^d$ is mapped to $\CM(\mu,\Circle)(x) = \alpha$ in the unit ball, the radius $\| \alpha \| \in [0,1]$ is interpreted as the typical centrality of $x$ with respect to distribution $\mu$. 

Let $p_0 \in \calP(\R^d)$ a reference measure, and $\mu,\nu \in \calP(\R^d)$ such that $p_0, \mu,\nu$ have densities. According to the previous definitions, a transport map $T \in \T(\mu,\nu)$ is \emph{$p_0-$quantile-preserving} (in the Monge-Kantorovich sense) if $\mu-$a.e.,
\begin{equation}
    \label{eq:p0-QP}
    \CM(\mu,p_0)(x) = \CM(\nu,p_0)(T(x)) \, .
\end{equation}
This identity implies that such a transport map is necessarily unique $\mu-$a.e..

\begin{theorem}[Quantile-preserving transport map]\label{thm:QPmaps}
    Let $p_0, \mu,\nu \in \calP(\R^d)$ with densities. The map
    \begin{equation}\label{eq:thm:QPmaps}
        \QP(\mu,\nu \, ; \, p_0) = \CM(p_0,\nu) \circ \CM(p_0,\mu)^{-1} 
    \end{equation} is a $p_0-$quantile-preserving transport map from $\mu$ to $\nu$.
    Moreover, this map is the unique $p_0-$quantile-preserving transport map in $\T(\mu,\nu)$ -- that is satisfying \eqref{eq:p0-QP} -- up to changing it on $\mu-$null measure sets.
    Next, $\QP(\mu,\nu \, ; \, p_0)$ will be defined as \emph{the} $p_0-$quantile-preserving ($p_0-$$\QP$) transport map from $\mu$ to $\nu$.
\end{theorem}

Note that $\CM(p_0,\mu)^{-1}$ is well defined $\mu-$a.e. via \cref{prop:OT_invertibility}. 
Before going further, it easy to check that $\QP(\mu,\nu \, ; \, p_0) \in \T(\mu,\nu)$: observe that if $X \sim \mu$, then by \cref{prop:OT_invertibility}, $\CM(p_0,\mu)^{-1}(X) \sim p_0$ thus $\QP(\mu,\nu \, ; \, p_0)(X) \sim \nu$.

In general, $\QP(\mu,\nu \, ; \, p_0)$ heavily depends on the reference measure $p_0$. However, when $d=1$, these quantile-preserving maps are independent of the reference measure as long as $\mu$ and $p_0$ have densities, and the following Lemma shows that the QP maps defined in \cref{thm:QPmaps} is consistent with the standard quantile-preserving map when $d=1$. 

\begin{lemma}[QP maps in 1D]\label{lem:QP_in_1D}
    When $d=1$, if $\mu$ and $p_0$ have densities, then
    $$\QP(\mu,\nu \, ; \, p_0) = F^{\dagger}_{\nu} \circ F_{\mu} \, .$$
\end{lemma}

Choosing $p_0=\Circle$ for the reference measure -- as in \cite{hallin2021distribution} -- has the advantage of this centrality interpretation. For other purposes however, one may want $p_0$ to be a product measure in order to represent quantile preservation over independent latent variables. In this case, a possible candidate is $p_0 = \square$, where $\square$ denotes the uniform distribution on $[0,1]^d$. This reference measure defines the \emph{hypercube quantile-preserving map}. If $\mu$ and $\nu$ themselves factorize, then $\QP(\mu,\nu \, ; \, p_0)$ does not depend on $p_0$, as long as $p_0$ factorizes -- see the proof of \cref{lem:productcase}.
Note that in general, as shown later \cref{prop:algebraic_classic}, $\QP(\mu,\nu \, ; \, p_0) \neq \QP(\mu,\nu \, ; \, q_0)$ even though $p_0$ and $q_
0$ both factorize.

\paragraph{QP maps and optimal transport.} As justified later, $\QP(\mu,\nu) \neq \CM(\mu,\nu)$, and thereby the QP map is not the optimal transport maps for the squared Euclidean cost. Nonetheless, the definition of $\QP(\mu,\nu)$ (\cref{thm:QPmaps}) and the invertibility property of $\CM(p_0,\nu)$ (\cref{prop:OT_invertibility}) shows that one can compute a QP maps by solving two optimal transport problems: the first from $\nu$ to $p_0$ and the second from $p_0$ to $\mu$.

\subsection{Triangular monotone transport map}

To motivate our third property, unlike the two first, we need to escape from dimension $d=1$. In many practical situations, when $d>1$, variables come with an intrinsic or imposed ordering. Consider for instance a sequential physical system or a time-dependent stochastic process where the state at time $t_k$ depends on past states $(t_1, \dots, t_{k-1})$. In such contexts, it is natural to seek a transport map which agrees with this structure by progressively pushing forward marginals and conditional distributions while preserving the order among coordinates, as notably argued by \cite{backhoff2017causal,bartl2024wasserstein}. The triangular monotone transport map, or Knothe–Rosenblatt (KR) map, offers precisely this: a transport that maps distributions $\mu$ and $\nu$ one coordinate at a time, without introducing backward dependencies between later and earlier variables, in a triangular, non-decreasing fashion.

\begin{definition}[Triangular maps, triangular monotonicity]
    A map $T: \R^d \to \R^d$ is \emph{triangular} if $T$ is of the form
    $$ T : (x_1, \ldots, x_d) \mapsto (T_1(x_1), T_2(x_1, x_2), \ldots, T_d(x_1, \ldots, x_d)) \, . $$ 
    $T$ is \emph{triangular monotone (TM)} if moreover, for all $1 \leq k \leq d$, all $(x_1, \ldots, x_{k-1}) \in \R^{k-1}$, $x_k \mapsto T_k(x_1, \ldots, x_{k-1}, x_k)$ is non-decreasing.
\end{definition}

Such maps were notably studied by \cite{rosenblatt1952remarks,knothe1957contributions,bogachev2005triangular}. In particular, \cite{rosenblatt1952remarks} proved that when $\mu$ and $\nu$ have densities, there is a $\mu-$a.e. \emph{unique} TM transport map from $\mu$ to $\nu$. Moreover, there is an explicit construction for it.

\begin{theorem}[\cite{rosenblatt1952remarks}]\label{thm:KR}
    Let $\mu,\nu \in \calP(\R^d)$. Assume that $\mu$ and $\nu$ have densities. Let $T_1(\cdot)$ be the CM transport map from $\mu_1$, the first marginal of $\mu$, to $\nu_1$, the first marginal of $\nu$. Recursively, for any $(x_1, \ldots, x_{k-1}) \in \R^{k-1}$, let
$ \mu_{k \, | \, (x_1, \ldots, x_{k-1})} $ denote the conditional distribution of the $k-$th coordinate under $\mu$ conditional to the $k-1$ first being $(x_1, \ldots, x_{k-1})$. Similarly define $ \nu_{k \, | \, (y_1, \ldots, y_{k-1})} $ for all $(y_1, \ldots, y_{k-1}) \in \R^{k-1}$.
Consider for almost every $(x_1, \ldots, x_{k-1}) \in \R^{k-1}$ the CM transport $T_k(x_1, \ldots, x_{k-1}, \cdot)$ given by:
$$ T_k(x_1, \ldots, x_{k-1}, \cdot) := \CM(\mu_{k \, | \, (x_1, \ldots, x_{k-1})}, \nu_{k \, | \, (T_1(x_1), \ldots, T_{k-1}(x_1,\ldots,x_{k-1}))}) \, .$$
This recursive procedure defines a triangular monotone transport map, called the \emph{Knothe-Rosenblatt (KR) transport map} from $\mu$ to $\nu$:
\[
\KR(\mu,\nu) : (x_1, \ldots, x_d) \mapsto (T_1(x_1), T_2(x_1, x_2), \ldots, T_d(x_1, \ldots, x_d)) \, .
\]
Moreover, the KR transport map is the unique triangular monotone transport map in $\T(\mu,\nu)$, up to changing it on $\mu-$null measure sets. 
\end{theorem}

Note that $\KR(\mu,\nu) \in \T(\mu,\nu)$ directly follows from the chain rule. Additionally $T = \KR(\mu,\nu)$ is easily invertible: since each $T_k$ is a one-dimensional monotone transformation conditioned on the previous variables, inversion can be performed sequentially. This makes the KR map especially useful in applications where one needs a fast and stable transport with a prescribed directionality or dependency structure.

\paragraph{KR maps and optimal transport}
$\KR$ maps are also closely related to optimal transport. For $\eps>0$, define the hierarchical $L^2$ norm $\| \cdot \|_{\eps}$ as follows:
$$ \| x \|^2_{\eps} := \sum_{i=1}^{d} \eps^i x_i^2 \, . $$

Assume additionally that $\mu$ and $\nu$ have finite second-order moments. Let $\OT_{\eps}(\mu,\nu)$ be the (well-defined) optimal solution with respect to this norm $\| \cdot \|_{\eps}$:
\begin{equation}\label{eq:OT_eps}
     \OT_{\eps}(\mu,\nu) := \argmin_{T \in \T(\mu,\nu)} \int \norm{x-T(x)}_{\eps}^2 \mathrm{d} \mu(x).
     \end{equation}
Note that $\OT_1(\mu,\nu) = \OT(\mu,\nu)$.\cite{carlier2010knothe} proved that the optimal transport map $\OT_{\eps}(\mu,\nu)$ converge in $L^2$ norm to $\KR(\mu,\nu)$ as $\eps \to 0$. In particular, one can compute an approximation of $\KR(\mu,\nu)$ as $\OT_{\eps}(\mu,\nu)$ for a small $\eps$. Other numerical approaches exist \citep{baptista2024representation}.

\begin{remark}[Imbalance in the hierarchical $L^2$ norm]
The increasing imbalance in the norm $\| x \|_{\eps}$ when $\eps \to 0$, giving way more importance to the first coordinate than the second, and so on, highlights the fact that the $\KR$ map reflects an inherent hierarchical structure over the coordinates of the random vectors in $\R^d$. The first marginals are matched first, followed by the second marginal conditioned on the first, and so on.
\end{remark}

\subsection{Results on collisions of transport maps}\label{sec:collision}
We hereafter give some sufficient conditions for the CM, QP, and KR transport maps to coincide. First, observe that the case where $d=1$ makes no difference between the three. 

\begin{lemma}[On the 1D case]
\label{lem:1Dcase}
    Note that when $d=1$, if $\mu, \nu$ have densities, then $\mu-$a.e.,
    $$ \CM(\mu,\nu) = \QP(\mu,\nu ; p_0) = \KR(\mu,\nu) = F^{\dagger}_{\nu} \circ F_{\mu}, $$ for any reference measure $p_0$ having a density. 
\end{lemma}

In fact, it is easily seen that these three transport maps are also identical in higher dimension $d \geq 1$ when $\mu,\nu$ are product measures.
\begin{lemma}[Case of product measures]
\label{lem:productcase}
    Assume $\mu, \nu$ are product measures $\mu = \otimes_{i=1}^{d} \mu_i$ and $\nu = \otimes_{i=1}^{d} \nu_i$ where the $\mu_i$ and $\nu_i$ have densities. Consider
    $$ T : x \mapsto \left(F^{\dagger}_{\nu_1} \circ F_{\mu_1}(x_1), \ldots, F^{\dagger}_{\nu_d} \circ F_{\mu_d}(x_d) \right) \, . $$ 
    Then $\mu-$a.e., 
    $$\CM(\mu,\nu) = \KR(\mu,\nu) = T\, .$$
    If moreover $p_0$ factorizes as $p_0 = \otimes_{i=1}^{d} p_{0,i}$ where the $p_{0,i}$ have densities, then $\mu-$a.e., 
    $$\CM(\mu,\nu) = \QP(\mu,\nu ; p_0) = \KR(\mu,\nu) = T\, .$$ 
\end{lemma}

A short comment on the last part of the last proof: note that in general, there is no reason why $\CM(p_0,\nu) \circ \CM(\mu,p_0)$ should itself be a CM map. As seen above, in the case where the measures are product measures, these maps are diagonal, and the cyclic monotonicity is preserved by composition. This is however an exception, as we will see next in \cref{prop:algebraic_classic} of \cref{sec:algebraic_prop}.

Beyond product measures, we conclude this subsection by providing a necessary and sufficient condition for the CM and KR maps to be equal. First off, we need a simple definition.

\begin{definition}[Diagonal non-decreasing maps]
    We say that $D : \R^d \to \R^d$ is \emph{diagonal non-decreasing} if $D$ is of the form $D(x)= (D_1(x_1), \ldots, D_d(x_d))$ with non-decreasing $D_i : \R \to \R$.
\end{definition}

If $D$ is diagonal non-decreasing, observe that $D$ is TM by definition. By Lebesgue differentiation theorem, the function $x \mapsto \int_{[0,x_1]} D_1(t_1)\rmd t_1 + \ldots + \int_{[0,x_d]} D_d(t_d) \rmd t_d$, convex since the $D_i$ are non-decreasing, is Lebesgue-a.e. differentiable and its gradient is $D$, Lebesgue-a.e. Thus, $D$ is also CM Lebesgue-a.e.

We just established that diagonal non-decreasing maps are both CM and triangular (thus also TM) Lebesgue-a.e. The next Proposition shows the converse. It is a fundamental result, which will be reused in next sections. 
For technical reasons, we need our result to hold if the CM and triangular maps agree $\mu-$a.e. This makes the proof less straightforward since one has to carefully deal with almost everywhere statements. 

\begin{proposition}[Triangular CM maps are the diagonal non-decreasing maps]
\label{prop:TM-CM_maps}
    Let $\mu \in \calP(\R^d)$ with a density. Let $C: \R^d \to \R^d$ be a CM map which we identify with $\CM(\mu, C\sharp\mu)$, and $T: \R^d \to \R^d$ be a triangular map. Assume that $\mu-$a.e., $C=T$. 
    Then, $C=T$ is $\mu-$a.e. diagonal non-decreasing.
\end{proposition}

We apply \cref{prop:TM-CM_maps} to get neccessary and sufficient conditions for CM and KR maps to coincide $\mu-$a.e.

\begin{corollary}[When $\CM(\mu,\nu)$ and $\KR(\mu,\nu)$ coincide]
   \label{cor:CM=KR}
   Let $\mu, \nu \in \calP(\R^d)$ having densities. The two propositions are equivalent:
\begin{itemize}
    \item[$(i)$] $\mu-$a.e., $\CM(\mu,\nu) = \KR(\mu,\nu)$.
    \item[$(ii)$] There exists a diagonal non-decreasing map $D : \R^d \to \R^d$ such that $\nu = D \sharp \mu$.
\end{itemize}
In this case, $\mu-$a.e., $\CM(\mu,\nu) = \KR(\mu,\nu)=D$.
\end{corollary}

\begin{figure}
    \centering
    \begin{subfigure}{0.3\textwidth}
    \includegraphics[width=\textwidth]{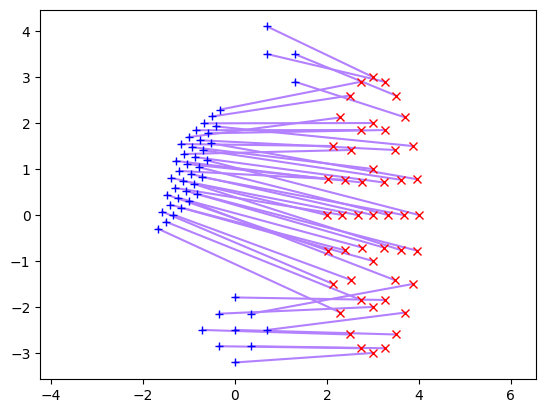}
    \caption{$\CM(\mu_n,\nu_n)$}
    \label{fig:CM}
    \end{subfigure}
    \hfill
    \begin{subfigure}{0.3\textwidth}
    \includegraphics[width=\textwidth]{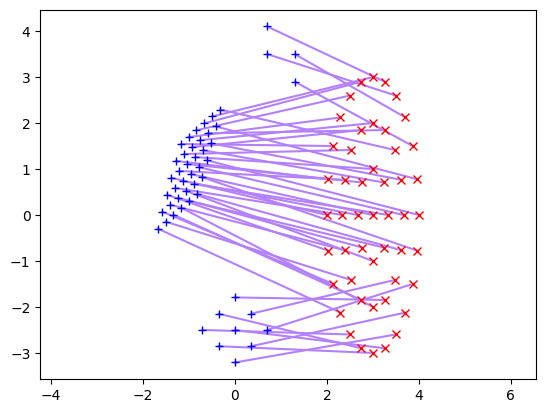}
    \caption{$\QP(\mu_n,\nu_n;\square_n)$}
    \label{fig:QP}
    \end{subfigure}
    \hfill
    \begin{subfigure}{0.3\textwidth}
    \includegraphics[width=\textwidth]{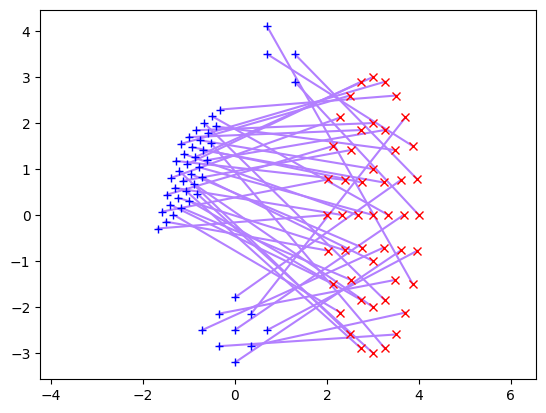}
    \caption{$\KR(\mu_n,\nu_n)$}
    \label{fig:KR}
    \end{subfigure}
    \caption{Three transport maps between the discrete measures $\mu_n$ (blue) and $\nu_n$ (red), with supports of cardinality $n=49$. The reference measure $\square_n$ corresponds to a uniform grid of $n=7^2$ points over $[0,1]^2$. The represented CM and QP transports correspond to the exact solutions obtained with optimal transport solvers; the represented KR transport is an approximation corresponding to the exact $\OT_\varepsilon(\mu_n,\nu_n)$ for $\varepsilon = 10^{-12}$. The code employs the \emph{Python Optimal Transport} package \citep{flamary2021pot}.}
    \label{fig:differences}
\end{figure}

Note that despite Lemmas \ref{lem:1Dcase}, \ref{lem:productcase} and \cref{cor:CM=KR}, as soon as $d \geq 2$ and $\mu,\nu$ are not product measures, the three transport maps are distinct in general. \Cref{fig:differences} illustrates this point between discrete 2D measures (which can be interpreted as discretizations of measures with densities).

\subsection{Algebraic properties of families of transport maps}\label{sec:algebraic_prop}

As a tool for our results and discussions in the next sections, we need to introduce general properties that one may consider for arbitrary families of transport maps between given marginals. They relate to the ones considered by \cite{dance2025counterfactual}. For the sake of consistency, we employ a similar terminology.

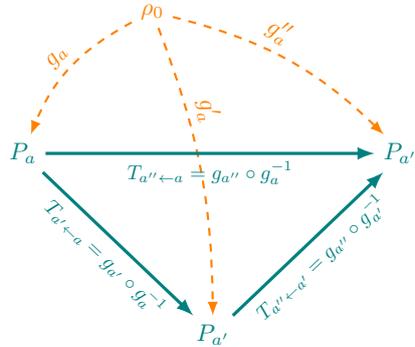
\begin{figure}[H]
    \centering
    \begin{tikzpicture}[scale=2, 
  thick,
  every node/.style={scale=0.9},
  ->, 
  >=latex,
  x={(1cm,0cm)}, 
  y={(0.5cm,0.8cm)}, 
  z={(0cm,1cm)}
]
\node[orange] (mu0) at (0.7,0.3,0.7) {$\rho_0$};
\node[teal] (Pa) at (0,0,0) {$P_a$};
\node[teal] (Pap) at (2,-1.5,0) {$P_{a'}$};
\node[teal] (Papp) at (2.5,0,0) {$P_{a''}$};
\draw[dashed, thick, orange] 
  (mu0) to[bend right=20] 
  node[midway, above, sloped] {$g_a$} 
  (Pa);
\draw[dashed, thick, orange] 
    (mu0) to[bend left=10] 
    node[pos=0.3, above, sloped] {$g_a'$} 
    (Pap);
\draw[dashed, thick, orange] 
  (mu0) to[bend left=15] 
  node[above left, sloped] {$g_a''$} 
  (Papp);
\draw[very thick,->, teal] (Pa) -- (Pap) node[midway,below, sloped, scale = 0.85, teal] 
  {$\displaystyle T_{a' \leftarrow a} = g_{a'} \circ g_a^{-1}$};
\draw[very thick,->, teal] (Pap) -- (Papp) node[midway,below, sloped, scale = 0.85, teal] 
  {$\displaystyle T_{a'' \leftarrow a'} = g_{a''} \circ g_{a'}^{-1}$};
\draw[very thick,->, teal] (Pa) to node[midway,below,sloped, scale = 0.85, teal]  
  {$\displaystyle T_{a'' \leftarrow a} = g_{a''} \circ g_a^{-1}$} (Papp);
\end{tikzpicture}
    \caption{Reference-based transport maps. Maps $g_a$ are themselves viewed as transport maps from some common reference measure $\rho_0$ to $P_a$.}
    \label{fig:ref_based_transports}
\end{figure}

\begin{definition}[Algebraic properties of transport maps]\label{def:algebraic}
Let $\A$ be a set, $(P_a)_{a \in \A}$ be a family of $\calP(\R^d)$, and $(T_{a' \leftarrow a})_{a,a' \in \A}$ be a collection of maps such that $T_{a' \leftarrow a} \in \T(P_a,P_{a'})$. We define three properties that $(T_{a' \leftarrow a})_{a,a' \in \A}$ may satisfy:
\begin{itemize}
    \item[$(i)$](Identity) for every $a \in \A$, $T_{a \leftarrow a} = \Id$;
    \item[$(ii)$](Path independence) for every $a,a',a'' \in \A$, $T_{a'' \leftarrow a} = T_{a'' \leftarrow a'} \circ T_{a' \leftarrow a}$;
    \item[$(iii)$](Inversion) for every $a,a' \in \A$, $T_{a' \leftarrow a}$ is invertible and $T^{-1}_{a' \leftarrow a} = T_{a \leftarrow a'}$.
\end{itemize}
\end{definition}
These properties automatically hold if the transport maps have a specific structure, illustrated in \cref{fig:ref_based_transports} and formalized below. 

\begin{lemma}[Reference-based transport maps]\label{lem:reference}
Let $\A$ be a set, $(P_a)_{a \in \A}$ be a family of $\calP(\R^d)$, and $(T_{a' \leftarrow a})_{a,a' \in \A}$ be a collection of maps such that $P_{a'} = T_{a' \leftarrow a} \sharp P_a$. If there exists a collection of injective functions $(g_a)_{a \in \A}$ such that $T_{a' \leftarrow a} = g_{a'} \circ g^{-1}_a$ for every $a,a' \in \A$, then $(T_{a' \leftarrow a})_{a,a' \in \A}$ meets all the properties from Definition~\ref{def:algebraic}.
\end{lemma}

Note that the notion of reference-based transport maps echoes the notion of \emph{coboundary maps} proposed by \cite{dance2025counterfactual}.

In line with our study of classical transport maps so far, the following proposition shows that $\CM, \QP$ and $\KR$ naturally meet some of these algebraic properties.
\begin{proposition}[Algebraic properties of classical transports]\label{prop:algebraic_classic}
Let $\A$ be a set and $(P_a)_{a \in \A}$ be a family of elements $\calP(\R^d)$ with densities. Then,
\begin{enumerate}
    \item[$(i)$] $(\CM(P_a,P_{a'}))_{a,a' \in \A}$ meet identity and inversion but not path independence in general;
    \item[$(ii)$] $(\QP(P_a,P_{a'};p_0))_{a,a' \in \A}$ (where $p_0 \in \calP(\R^d)$ has a density) meet identity, path independence, and inversion;
    \item[$(iii)$] $(\KR(P_a,P_{a'}))_{a,a' \in \A}$ meet identity, path independence, and inversion.
\end{enumerate}
\end{proposition}

\begin{remark}[On CM maps satisfying path independence]
    A family of CM transport maps may satisfy path independence for \emph{specific} collections of marginals. This is the case for instance if the collection of maps is diagonal non decreasing, or radial, that is with CM maps of the form $x \mapsto \frac{x}{\|x \|} f_{a,a'}(\|x \|)$, where the $f_{a,a'}$ are continuous, non decreasing, and satisfy $f_{a,a'}(0)=0$. See \citep[Section 4.1]{boissard2015distribution} for more exhaustive results and details. Note that path independence always holds whenever $\abs{\A} = 2$.
    
    We also emphasize that the fact that CM maps are not stable under composition in general explains why $\QP(\mu,\nu) \neq \CM(\mu,\nu)$. Moreover, this lack of algebraic property will play a critical role in understanding the applicability of CM maps for counterfactual reasoning: a topic we address next.
\end{remark}

To conclude this section, we summarize the main properties of the three transports in the following table.
\begin{table}[h]
    \centering
    \resizebox{\columnwidth}{!}{
    \begin{tabular}{|c|c|c|c|c|}
         \hline
         Transport & Monotonicity & Quantile invariance & Optimality & Algebraic properties \\
         \hline
         $\CM$ & CM & -- & for $\| \cdot \|^2$ & identity, inversion \\
         $\QP$ & -- & Multivariate quantiles & -- & identity, path independence, inversion \\
         $\KR$ & TM & Univariate conditional quantiles & for $\| \cdot \|_\eps^2$, when $\eps \to 0$ & identity, path independence, inversion \\
         \hline
    \end{tabular}
    }
    \caption{Three canonical transport maps and their properties.}
    \label{tab:transports_comparison}
\end{table}

\section{Counterfactual reasoning: a mass-transportation viewpoint}\label{sec:counterfactual}

The previous section defined transport maps between preassigned marginals and presented classical choices of such maps in generic applications from statistics and machine learning. Interestingly, transport maps also emerge in counterfactual reasoning, the finest form of causal inference \citep{pearl2018book}, which answers questions of the form \say{Had Alice taken aspirin, would have her headache been cured?} \citep{lewis2013counterfactuals}.

In this section, we present counterfactual reasoning with Pearl's causal framework and explain how this can be reframed as a mass-transportation problem. Section~\ref{sec:causal_prelim} provides a general background on causal modeling while Section~\ref{sec:setup} presents a specific setting that relates counterfactuals to transport maps.

\subsection{Preliminaries}\label{sec:causal_prelim}

Pearl's causal framework \citep{pearl2009causality} deals with two types of causal models: causal graphical models and structural causal models. Crucially, only structural causal models allow full counterfactual inference.

A structural causal model can be seen as a data-generating process, which specifies how exogenous variables (representing latent causes) produce endogenous variables (encoding observational phenomena) via cause-effect relationships. 
\begin{definition}[Structural causal model]\label{def:scm}
Let $\I$ be finite index set. A \emph{structural causal model} (SCM) $\M$ with \emph{endogenous variables} $\I$ is a tuple $(\V,\G,\U,Q,f)$ where
\begin{itemize}
    \item $\V := \times_{i \in \I} \V_i$ and $\U := \times_{i \in \I} \U_i$ are products of Borel subsets of $\R$, respectively called the \emph{space of endogenous values} and \emph{space of exogenous values};
    \item $\G$ is a directed graph with nodes $\I$ and parental function $\pa : \I \to 2^{\I}$, called the \emph{causal graph};
    \item $Q := \otimes_{i \in \I} Q_i$ is a product measure over $\U$ such that $Q_i \in \calP(\U_i)$ for every $i \in \I$, called the \emph{exogenous distribution};
    \item $f := (f_i)_{i \in \I}$ is a collection a measurable functions $f_i : \V_{\pa(i)} \times \U_i \to \V_i$ for every $i \in \I$, called the \emph{causal mechanism}.
\end{itemize}
A pair $(V,U)$ of random vectors $V : \Omega \to \V$ and $U : \Omega \to \U$ is a \emph{solution} of $\M$ if $\law{U} = Q$ and $V$ solves the \emph{structural equations}, that is, $V_i = f_i(V_{\pa(i)},U_i)$ for every $i \in \I$.\footnote{In the examples to come, we often specify an SCM via only structural equations for the sake of simplicity.}
\end{definition}
This definition has two particularities compared to other common definitions from the literature. It is notably more general than that from \cite[Definition 6.2]{peters2017elements}, and more restrictive than that from \cite[Definition 2.1]{bongers2021foundations}. First, the graph $\G$ can be \emph{cyclic}. Second, every endogenous variable is caused by at most one exogenous variables and the exogenous variables are mutually independent. Throughout, we write $\pa(I) = \cup_{i \in I} \pa(i)$ for any $I \subseteq \I$. Additionally, for every $i \in \I$, we denote by $\an(i)$ the set of ancestors of $i$ in $\G$---which contains $i$.

An SCM serves to formalize associations that standard probability calculus cannot describe through the notion of intervention. An intervention represents the action of forcing certain variables to take predefined (possibly contrary-to-fact) values while keeping the rest of the mechanism untouched.

\begin{definition}[Perfect intervention]\label{def:do}
Let $\I$ be a finite index set and $\M := (\V,\G,\U,Q,f)$ be an SCM over $\I$. For any $I \subseteq \I$ and any $v^\star_I \in \V_I$, the operation $\doint(I,v^\star_I)$ maps $\M$ to the SCM $\M_{\doint(I,v^\star_I)} := (\V,\tilde{\G},\U,Q,\tilde{f})$, where $\tilde{\G}$ is obtained by removing all incoming arrows to $I$ and
\[
    \tilde{f_i}(v_{\pa(i)},u_i) := \begin{cases}
                    v^\star_i \text{ if } i \in I,\\
                    f_i(v_{\pa(i)},u_i) \text{ otherwise}.
                   \end{cases}  
\]
We call such an operation a \emph{perfect intervention}. 
\end{definition}

In general, because $\G$ can be cyclic, an SCM and its postintervention counterparts do not necessarily admit solutions. This motivates a definition of solvability, adapted from \citep[Definition 3.3]{bongers2021foundations}.
\begin{definition}[Unique solvability of SCMs]\label{def:solvability}
Let $\I$ be a finite index set and $\M := (\V,\G,\U,Q,f)$ be an SCM over $\I$. An SCM is \emph{uniquely solvable with respect to $I \subseteq \I$} if and only if there exists a measurable mapping $S_I : \V_{\pa(I) \setminus I} \times \U_I \to \V_I$ such that for $Q$-almost every $u \in \U$ and every $v \in \V$,
\[
    v_I = S_I(v_{\pa(I) \setminus I}, u_I) \iff v_i = f_i(v_{\pa(i)},u_i) \text{ for every } i \in I.
\]
We call $S_I$ the solution map of $\M$ relative to $I$, and say that $\M$ is uniquely solvable if it is uniquely solvable with respect to $\I$. Note that $S_\I$ is a function of the exogenous variables only.
\end{definition}
By definition, the solution map relative to a given subset of variables is unique up to sets of probability measure zero. For simplicity, we will always pick a single representation among all the undistinguishable possibilities and refer to it as \emph{the} solution map. We recall that solution with respect to $I$ does not imply solvability with respect to $J \subseteq I$ in general \citep[Example 2.8]{bongers2021foundations}.

Following \citep[Theorem 3.6]{bongers2021foundations}, if an SCM is uniquely solvable, then it admits a solution $(V,U)$. Furthermore, for any solution $(V,U)$, the law of $V$ is unique, determined by $\law{V} = S_{\I} \sharp Q$. We propose to illustrate the three above definitions in an example below.

\begin{example}[Interventions and solution maps]\label{ex:solution_maps}
We consider a medical example inspired by studies on the genetics of smoking \citep{davies2009genetics,mackillop2010role}. Let $X_1$ quantify the expression of a gene, let $A$ indicate whether the person smokes, and let $X_2$ encode a score of health. We suppose that these three random variables are governed by an SCM $\M := (\V,\G,\U,Q,f)$ corresponding to the following assignments,
\[
\begin{cases}
    X_1 = U_1\\
    A = \indic{X_1+U_{\tta}>0}\\
    X_2 = \alpha X_1 + \beta A + U_2\\
\end{cases},
\]
where $(U_1,U_{\tta},U_2) \sim Q$. We emphasize that, at this stage, there is no justification that $(X_1,A,X_2)$ is well-defined given $(U_1,U_{\tta},U_2)$, namely, that $\M$ is uniquely solvable. Essentially, the question of the unique solvability of $\M$ is equivalent to the question of solving the above system of equations in a measurable way, where the unknown variable is $(X_1,A,X_2)$ while $(U_1,U_{\tta},U_2)$ are known parameters. The triangular form of this system, which results from the acyclicity of $\G$, enables us to write:
\[
\begin{cases}
    X_1 = U_1\\
    A = \indic{U_1+U_{\tta}>0}\\
    X_2 = \alpha U_1 + \beta \indic{U_1+U_{\tta}>0} + U_2\\
\end{cases}.
\]
Therefore, $\M$ is uniquely solvable with solution map
\[
    S(u_1,u_{\tta},u_2) = \begin{pmatrix}
        u_1\\
        \indic{u_1+u_{\tta}>0}\\
        \alpha u_1 + \beta \indic{u_1+u_{\tta}>0} + u_2\\
    \end{pmatrix}.
\]
The model $\M$ is also uniquely solvable with respect to $\{1, 2\}$, since the same system can be solved with $(X_1,X_2)$ as unknown variable and $(A,U_1,U_2)$ as known parameters:
\[
\begin{cases}
    X_1 = U_1\\
    X_2 = \alpha X_1 + \beta A + U_2\\
\end{cases}.
\]
The corresponding solution map is given by
\[
    S_{\ttx}(a',u_1,u_2) = \begin{pmatrix}
        u_1\\
        \alpha u_1 + \beta a' + u_2\\
    \end{pmatrix}.
\]

Next, we apply the perfect intervention $\doint(\tta,a)$ on $\M$, where $a$ is a fixed admissible value. This produces the SCM associated to
\[
\begin{cases}
    X_{a,1} = U_1\\
    A_a = a\\
    X_{a,2} = \alpha X_{a,1} + \beta A_a + U_2\\
\end{cases},
\]
where $(U_1,U_{\tta},U_2) \sim Q$. Again, $\M_a$ is uniquely solvable relative to $\{1, \tta, 2\}$ and $\{1, 2\}$. The corresponding solution maps are, respectively, given by
\[
    S^{(a)}(u_1,u_{\tta},u_2) = \begin{pmatrix}
        u_1\\
        a\\
        \alpha u_1 + \beta a + u_2\\
    \end{pmatrix}
    = \begin{pmatrix}
        S_{\ttx}(a,u_{\tta},u_2)_{1}\\
        a\\
        S_{\ttx}(a,u_{\tta},u_2)_{2}\\
    \end{pmatrix}
\]
and
\[
    S^{(a)}_{\ttx}(a',u_1,u_2) = \begin{pmatrix}
        u_1\\
        \alpha u_1 + \beta a + u_2\\
    \end{pmatrix} = \begin{pmatrix}
        S_{\ttx}(a,u_{\tta},u_2)_{1}\\
        S_{\ttx}(a,u_{\tta},u_2)_{2}\\
    \end{pmatrix}.
\]
We underline that $S^{(a)}_{\ttx}$ is a constant function in its first argument $a'$. The highlighted connection between $S^{(a)}$, $S^{(a)}_{\ttx}$, and $S_{\ttx}$ will be made more formal in the next subsection.
\end{example}

We now advance to counterfactual reasoning.

\subsection{Problem setup}\label{sec:setup}

Counterfactual reasoning can be defined as thinking about hypothetical outcomes in alternative worlds where some circumstances changed from what factually happened. It corresponds to the sharpest level of causal inference, as it generally cannot be falsified even via randomized experiments. For illustration, considers a clinical trial, where a medical treatment is blindly assigned to some individuals, leading to exchangeable control and treated groups. As such, one can assess the general causal effect of the treatment onto recovery by testing the statistical difference between the groups' outcomes. However, one cannot fully determine the singular (or individual) causal effect, namely the effect on a single individual. Say, for instance, that Alice was in the treated group and actually recovered. No experiment can prove or disprove that Alice would not have recovered had she been in the control group. Therefore, the evaluation of counterfactual statements rests on conviction, on choices. Despite their unfalsifiability, counterfactuals underpin fundamental concepts and applications (as mentioned in the introduction) at the core of actual research in causality \citep{kusner2017counterfactual,mesnard2021counterfactual,richens2022counterfactual,mueller2023personalized,sarvet2025perspectives}.

In the following, we explain how SCMs allow to formalize such choices, and thereby enable analysts to carry out counterfactual inference. We focus on a setting where the counterfactuals are said to be \emph{deterministic}. Interestingly, this makes counterfactual pairs (like the actual Alice and her counterfactual counterpart) stem from transport maps between the factual and contrary-to-fact worlds (like the treated and control groups). 

\subsubsection{Causal model and deterministic setting}

We introduce further formalism to describe counterfactuals with SCMs. Let $\I := \{ 1,\ldots, d, \tta \}$ be an index set of endogenous variables. We consider SCMs $\M := (\V,\G,\U,Q,f)$ over $\I$, where $\tta$ can have any position in $\G$. Throughout the rest of the paper, we employ the following notation: $\A := \V_{\tta}$, $\X := \times^d_{i=1} \V_i$, $\U_{\ttx} := \times^d_{i=1} \U_i$, and $Q_{\ttx} := \otimes^d_{i=1} Q_i$. To keep the subscripts simple and intuitive, we index any endogenous tuple $v \in \V$ as $v = (x_1,\ldots,x_d,a)$, with $x \in \X$ and $a \in \A$. Similarly, we index any exogenous tuple $u \in \U$ as $u = (u_1,\ldots,u_d,u_{\tta})$, with $u_{\ttx} := (u_1,\ldots,u_d) \in \U_{\ttx}$ and $u_{\tta} \in \U_{\tta}$. We tackle the problem of reasoning about potential outcomes of $\ttx$ had $\tta$ taken contrary-to-fact values in $\A$. For every $a \in \A$, we denote by $\M_a$ the result of $\doint(\tta,a)$ on $\M$, which describes an alternative world where $\tta$ is forced to take the value $a$ while keeping the rest of the data-generating process equal. Hereafter, we introduce a derivative of $\M_a$ that will play a pivotal role in our main assumptions and results.

\begin{proposition}[Subsolution map]\label{prop:submap}
Let $\M := (\V,\G,\U,Q,f)$ be an SCM over $\{ 1,\ldots, d, \tta \}$, using the same notation as above. If $\M$ is uniquely solvable with respect to $\{1,\ldots, d\}$, then, for every $a \in \A$,
\begin{enumerate}
    \item[$(i)$] $\M_a$ is uniquely solvable with respect to $\{ 1,\ldots, d, \tta \}$,
    \item[$(ii)$] the measurable mapping $g_a : \U_{\ttx} \to \X, u_{\ttx} \mapsto S_{\ttx}(a_{\pa([d])},u_{\ttx})$ is well defined and satisfies, for every solution $(X_a,a,U)$ of $\M_a$, $X_a = g_a(U_{\ttx})$.
\end{enumerate}
For every $a \in \A$, we call $g_a$ the subsolution map of $\M_a$. 
\end{proposition}

Next, we suppose that $\M$ meet the conditions below to ensure the existence of counterfactuals and to guarantee that they are deterministic. We will properly define later what \say{deterministic} means.

\begin{assumption}\label{hyp:scms}
$\M$ is an SCM over $\{1,\ldots, d, \tta\}$ such that
\begin{itemize}
    \item[$(i)$] $\M$ is uniquely solvable with respect to $\{1,\ldots, d, \tta\}$ and $\{1,\ldots, d\}$;
    \item[$(ii)$] for every $a \in \A$, the subsolution map $g_a$ of $\M_a$ is injective.
\end{itemize}
We denote by $\SCM$ the set of SCMs meeting this assumption.
\end{assumption}

The first part of this assumption automatically holds for acyclic SCMs. More specifically, acyclic SCMs and their postintervention counterparts are uniquely solvable with respect to any subset of variables. Solution maps are recursively obtained by unrolling the structural equations along the topological order induced by the graph, as illustrated by \cref{ex:solution_maps} and proved later in \cref{prop:acyclic}. As such, we introduce the solvability conditions in \cref{hyp:scms} precisely to address possibly cyclic models. In this sense, these conditions share similarities with the definition of simple SCMs \citep[Section 8]{bongers2021foundations}. The key difference lies in the fact that we do not assume unique solvability with respect to \emph{any} subset of variables; only specific subsets. The second part of this assumption is standard in order to render the entailed counterfactuals \emph{deterministic}. We refer to notably \citep{karimi2021algorithmic,delara2024transport,chen2025exogenous} for an analogous condition. Section \ref{sec:counterfactual_maps} details this point.

Under \cref{hyp:scms}, $\M$ and $\M_a$ for every $a \in \A$ are uniquely solvable, and thereby admit solutions. Writing $(X,A,U)$ for a solution of $\M$, we denote by $(X_a,a,U)$ a solution of $\M_a$ \emph{with the same exogenous noise $U$}. To clarify the subscripts, $X_i$ refers to the $i$th component of $X$ while $X_{a,i}$ refers to the $i$th component of $X_a$. Then, $X_a = g_a(U_{\ttx})$ for every $a \in \A$. Finally, for every $a \in \A$, we refer to $P_a := \law{X_a} = g_a \sharp Q_{\ttx}$ as the \emph{interventional marginal} had $\tta$ been equal to $a$.\footnote{We point out that, in \cref{sec:meet}, $(P_a)_{a \in \A}$ may as well denote the interventional marginals of some SCM (as here in \cref{sec:counterfactual}) or an arbitrary collection of marginals (as before in \cref{sec:transport}). The role of $(P_a)_{a \in \A}$ will be explicitly stated in each result.} 

Applying this formalism to the medical illustration from \cref{ex:solution_maps}, the interventional marginals $P_0$ and $P_1$ correspond to the outcomes of, respectively, the control and treated group. They represent falsifiable quantities and allow to compute the causal effect of $A$ onto $X$ at the group scale. In \cref{ex:solution_maps}, a cross-world solution $((X_{0,1},X_{0,2}),(X_{1,1},X_{1,2}))$ describes single individuals across distinct realities, entangled by the same exogenous noise $U$. It induces a joint probability distribution between $P_0$ and $P_1$ called a \emph{counterfactual coupling}. Such an object is determined only by the SCM; it does not relate to experimental quantities.

\subsubsection{Counterfactual maps}\label{sec:counterfactual_maps}

In order to make connections with Section~\ref{sec:transport}, we deliberately considered a setting where counterfactual couplings are deterministic, that is, related by invertible transport maps. To prove this point, note that $U_{\ttx} = g^{-1}_a(X_a)$, according to \cref{hyp:scms}. Applying $g_{a'}$ both sides of this equality gives $X_{a'} = (g_{a'} \circ g^{-1}_a)(X_a)$. This defines a deterministic relation between $X_a$ and $X_{a'}$.

\begin{definition}[Counterfactual map]
For any $\M \in \SCM$ and every $a,a' \in \A$, we call
\[
    C_{a' \leftarrow a} := g_{a'} \circ g^{-1}_a
\]
the \emph{counterfactual map of $\M$ had $\tta$ been equal to $a'$ instead of $a$}.
\end{definition}

The interpretation is the following. The function $g_{a'}$ determines an outcome in the world where $\tta$ is set to value $a'$ from a realization of the exogenous distribution, while the function $g^{-1}_a$ recovers the exogenous input that produced an outcome in the world where $\tta$ is set to value $a$. As such, a counterfactual map forms pairs of outcomes across parallel worlds through common exogenous values. Crucially, $P_{a'} = {C_{a' \leftarrow a}} \sharp P_a$: at the \emph{counterfactual}-level of causation, an SCM induces a transport map between \emph{interventional} marginals $P_a$ and $P_{a'}$. We emphasize that---because the counterfactual coupling $(X_a,X_{a'})$ is deterministic---the computation of counterfactual counterparts does not involve the exogenous distribution $Q_{\ttx}$. For illustration, we continue \cref{ex:solution_maps}.

\begin{example}[Counterfactuals]\label{ex:counterfactuals}
For every $a \in \A$, the subsolution map $g_a$ of $\M_a$ is by definition
\[
    g_a(u_1,u_2) = \begin{pmatrix}
        u_1\\
        \alpha u_1 + \beta a + u_2\\
    \end{pmatrix},
\]
while its inverse is given by
\[
    g^{-1}_a(x_1,x_2) = \begin{pmatrix}
        x_1\\
        x_2 - \alpha x_1 - \beta a\\
    \end{pmatrix}.
\]
In this specific example, observe that the expression of $g^{-1}_a$ can readily be obtained by isolating the exogenous variables in the system of structural equations, without inversing $g_a$ explicitly. Using these two formulas, for every $a,a' \in \A$, the counterfactual map is expressed as
\[
    C_{a' \leftarrow a}(x_1,x_2) = \begin{pmatrix}
        x_1\\
        \beta(a'-a) + x_2\\
    \end{pmatrix}.
\]
Another way to recover the counterfactual maps is to consider a solution $(X_a,a,U)$ of $\M_a$ and a solution $(X_{a'},a',U)$, with the same exogenous noise $U$. They satisfy
\[
\begin{cases}
    X_{a,1} = U_1\\
    X_{a,2} = \alpha X_{a,1} + \beta a + U_2\\
\end{cases},
\]
and
\[
\begin{cases}
    X_{a',1} = U_1\\
    X_{a',2} = \alpha X_{a',1} + \beta a' + U_2\\
\end{cases}.
\]
Then, simply note that 
\[
\begin{cases}
    X_{a',1} = X_{a,1}\\
    X_{a',2} = \beta(a'-a) + X_{a,2}\\
\end{cases}.
\]
This identifies $C_{a' \leftarrow a}$ via $X_{a'} = C_{a' \leftarrow a}(X_a)$.
\end{example}

All in all, under some assumptions, an SCM entails a family of transport maps, that characterizes counterfactual beliefs. The next section addresses the natural question of which SCMs entail which families of counterfactual maps. To properly conduct this discussion, we need some fundamental properties of counterfactual maps. First, note that the counterfactual maps are reference-based as defined in \cref{lem:reference}, leading to the following result.
\begin{proposition}[Algebraic properties of counterfactual maps]\label{prop:algebraic_ctf}
Let $\M \in \SCM$. Then, $(C_{a' \leftarrow a})_{a,a' \in \A}$ meet all the properties from \cref{def:algebraic}, that is, identity, path independence and inversion.
\end{proposition}

We underline that \cite{dance2025counterfactual} were the first to highlight the algebraic constraints on counterfactual maps.

Second, again due to their composite forms, counterfactual maps are stable under specific reparameterizations of the latent SCM, namely, under changes of exogenous distributions.

\begin{proposition}[Exogenous reparameterization]\label{prop:reparameterization}
Let $\M \in \SCM$ and $(\phi_i)_{i \in \I}$ be a collection of bijective measurable maps $\phi_i : \widetilde{\U}_i \to \U_i$, where $\I := [d] \cup \{ \tta \}$. Define $\widetilde{\M} := (\V,\G,\widetilde{\U},\widetilde{Q},\widetilde{f})$ such that, for every $i \in \I$, $\widetilde{Q}_i = \phi^{-1}_i \sharp Q_i$ and $\widetilde{f}_i = f_i \circ (\Id \times \phi_i)$. Then, $\widetilde{\M} \in \SCM$, and write for every $a,a' \in \A$, $\widetilde{g}_a$ and $\widetilde{C}_{a' \leftarrow a}$ for, respectively, the subsolution map of $\widetilde{\M}_a$ and the counterfactual map from $a$ to $a'$ of $\widetilde{\M}$. It holds that, $\widetilde{g}_a = g_a \circ (\times^d_{i=1} \phi_i)$ and $\widetilde{C}_{a' \leftarrow a} = C_{a' \leftarrow a}$. 
\end{proposition}

\subsection{Two specific cases}

\Cref{ex:solution_maps} and \cref{ex:counterfactuals} detail how to derive the solution maps and the counterfactual maps from a simple SCM, where the graph is acyclic and the structural equations on a subset of variables are linear additive. In what follows, we generalize this analysis to two classes of models, acyclic SCMs and partially linear additive SCMs, which will play a role in \cref{sec:meet}.

\subsubsection{Acyclic SCMs}

We say that an SCM is \emph{acyclic} if its graph is a \emph{directed acyclic graph} (DAG). A DAG induces a topological order over its nodes. Throughout the rest of the paper, we always assume that the variables $\{1,\ldots,d\}$ of any acyclic SCM over $\{1,\ldots,d,\tta\}$ are sorted according to this order. We recall that $\tta$ can be positioned anywhere in the graph. This can be done without loss of generality, in the sense that it suffices to apply a permutation on the variables. As aforementioned, acyclic models are uniquely solvable with respect to any subset of variables. We consider acyclic SCMs with noise-invertible causal mechanisms, which we encapsulate in the following assumption.

\begin{assumption}\label{hyp:acy}
$\M$ is an SCM over $\{1,\ldots,d,\tta\}$ such that,
\begin{itemize}
    \item[$(i)$] $\M$ is acyclic;
    \item[$(ii)$] for every $i \in [d], x \in \X, a \in \A$, the function $f_i(a_{\pa(i)},x_{\pa(i)},\cdot)$ is injective.
\end{itemize}
We denote by $\SCM_{acy}$ the set of SCMs meeting this assumption. Moreover, for any DAG $\G$ we write $\SCM^{\G}_{acy}$ for the subset of $\SCM_{acy}$ whose elements all have $\G$ as causal graph.
\end{assumption} 

In acyclic SCMs with noise-invertible causal mechanisms, both the subsolution maps and their inverses take an explicit and intuitive form.

\begin{proposition}[Solvability of acyclic SCMs]\label{prop:acyclic}
If $\M \in \SCM_{acy}$, then $\M$ is uniquely solvable with respect to any subset of variables and is such that $g_a$ is injective for every $a \in \A$. In other words, $\SCM_{acy} \subseteq \SCM$.

More specifically, for every $\M \in \SCM_{acy}$ and every $a \in \A, i \in [d]$, there exist measurable maps $\psi_i, \tilde{\psi}_i$ such that:
\begin{enumerate}
    \item[$(i)$] for all $u_{\ttx} \in \U_{\ttx}$, $g_a(u_{\ttx})_i = f_i(a_{\pa(i)},\psi_i(a,u_{\an(i) \setminus \{i\}}),u_i)$,
    \item[$(ii)$]
    for all $x \in \X$, $g^{-1}_a(x)_i = f_i(a_{\pa(i)},\tilde{\psi}_i(a,x_{\an(i) \setminus \{i\}}),\cdot)^{-1}(x_i)$.
\end{enumerate}
In particular, for every $a,a' \in \A$, $g_a$, $g^{-1}_a$, and $C_{a' \leftarrow a}$ are lower triangular maps.
\end{proposition}

While \cref{prop:acyclic} is notationally involved, it follows an intuitive principle: one can always solve a triangular system of structural equations by recursively unrolling the equations of the SCM along the topological order. Notably, the $i-$th component of a solution depends only on the exogenous components associated to $\an(i)$.

Note that \cref{prop:acyclic} specifies \cref{prop:submap} by clarifying the structure of the subsolution map in the acyclic case. In particular, this proves that $\SCM_{acy} \subseteq \SCM$. Naturally, \cref{hyp:acy} serves to consider simpler settings than those captured by \cref{hyp:scms}, . The second part of the assumption notably ensures that the subsolution maps are invertible, so that the entailed counterfactuals are deterministic. Such SCMs are sometimes referred to as \emph{bijective SCMs} \citep{nasr2023counterfactual,dance2025counterfactual}.

\subsubsection{Linear additive SCMs}

We turn to possibly cyclic models, but with a particular parametric form: (partially) linear models, which are widely considered in causality research and applications.

\begin{definition}[Partially linear-additive model]
An SCM $\M := (\V,\G,\U,Q,f)$ over $d$ variables is a \emph{linear additive model} relative to $I \subseteq [d]$ if there exist a matrix $M := (M_{i,j})_{i,j \in [d]} \in \R^{d \times d}$ satisfying $M_{i,j} = 0 \iff j \notin \pa(i)$ and a vector $b := (b_i)^d_{i=1} \in \R^d$, such that for every $i \in I$
\[
    f_i(v_{\pa(i)},u_i) =  \sum^d_{j=1} M_{i,j} v_j + b_i + u_i.
\]
for every $i \in [d]$, $u \in \U$, and $v \in \V$.
\end{definition}

The interest of these models comes from their simplicity. Notably, one can derive closed-forms for their counterfactual maps. To do so, we first need the expression of their solution maps.

\begin{lemma}[Unique solvability of linear additive models]\label{lem:linear}
Let $\M$ be a linear additive SCM relative to $I \subseteq [d]$. Then, $\M$ is uniquely solvable with respect to $I$ if and only if the submatrix $\Id - M_{I,I}$ is invertible. In this case, the solution map is given by 
\[S_I(v_{\pa(I) \setminus I},u_I) = (\Id - M_{I,I})^{-1}(M_{I,\pa(I) \setminus I} v_{\pa(I) \setminus I} + b_I + u_I).
\]
\end{lemma}

One can check that $\Id - M_{I,I}$ is always invertible in the acyclic case (since $M_{I,I}$ is nilpotent), which is consistent with the fact that acyclic SCMs are uniquely solvable. Crucially, the invertibility condition also holds for some cyclic models. On this basis, we determine the counterfactual maps.

\begin{proposition}[Counterfactuals of linear additive models]\label{prop:ctf_linear}
Let $\M \in \SCM$ be a linear additive model over $\{1,\ldots,d,\tta\}$. Then, there exist a matrix $M^{\ttx} \in \R^{d \times d}$ satisfying $M^{\ttx}_{i,j} = 0 \iff j \notin \pa(i)$ and $\Id - M^{\ttx}$ is invertible, a vector $m^{\ttx} \in \R^d$ meeting  $m^{\ttx}_i = 0 \iff \tta \notin \pa(i)$, along with $b^{\ttx} \in \R^d$, such that, for every $a \in \A$ and $u_{\ttx} \in \U_{\ttx}$,
\[
    g_a(u_{\ttx}) = (\Id - M^{\ttx})^{-1}(m^{\ttx} a + b^{\ttx} + u_{\ttx}).
\]
Therefore, for every $a',a \in \A$
\[
    C_{a' \leftarrow a}(x) = x + (\Id - M^{\ttx})^{-1} m^{\ttx} (a' - a) \, .
\]
\end{proposition}

To summarize, linear additive SCMs---even cyclic ones---generate counterfactuals via mere translations between the interventional marginals. Interestingly, translations are monotone functions in the sense of some definitions studied in \cref{sec:transport}. We dedicate the thorough analysis of the connections between the classical transport maps from \cref{sec:transport} and the counterfactual maps from \cref{sec:counterfactual} to the section below.

\section{When counterfactual maps meet classical transports}\label{sec:meet}

In this section, we apply the unified transport-map formalism to analyze possible links between the classical transport maps from Section~\ref{sec:transport} and the counterfactual maps from Section~\ref{sec:counterfactual}. Concretely, we ask: under which conditions on the SCM $\M$ are the counterfactual maps $(C_{a' \leftarrow a})_{a,a' \in \A}$ optimal-transport maps, or quantile-preserving maps, or Knothe-Rosenblatt transports?

\subsection{From counterfactuals to solution maps}
A counterfactual map is related to the subsolution maps via $C_{a' \leftarrow a} = g_{a'} \circ g^{-1}_a$. Therefore, any property on the counterfactual maps may constrain the subsolution maps. Conversely, any property on the subsolution maps may constrain the counterfactual maps. We propose two results shedding light on this relationship: one for general (possibly cyclic) SCMs; the other for acyclic SCMs. For the purpose of this analysis, we introduce notions of \emph{co}monotone functions, which relate to the notions of monotonicity discussed in \cref{sec:transport}.

\begin{definition}[Notions of comonotonicity]
We introduce three definitions.
\begin{itemize}
    \item[$(i)$] Two functions $f,g : \R \to \R$ are \emph{comonotone} if, for every $x,y \in \R$,
    \[
        (f(x)-f(y))(g(x)-g(y)) \geq 0.
    \]
    \item[$(ii)$] Two functions $f,g : \R^d \to \R^d$ are \emph{cyclically comonotone} if, for any $m \geq 1$ and any $x^{(1)},x^{(2)},\ldots,x^{(m)} \in \R^d$,
    \[
        \sum^m_{i=1} \langle f(x^{(i)}), g(x^{(i)}) - g(x^{(i+1)}) \rangle \geq 0,
    \]
    where $x_{m+1}=x_1$.
    \item[$(iii)$] Two functions $f,g : \R^d \to \R^d$ are \emph{diagonally comonotone} if, for every $i \in [d]$, every $x_1,\ldots,x_{i-1},x_{i+1},\ldots,x_d \in \R$, and every $x_i,x'_i \in \R$,
    \begin{multline*}
        (f_i(x_1,\ldots,x_i,\ldots,x_d)-f_i(x_1,\ldots,x'_i,\ldots,x_d))\\(g_i(x_1,\ldots,x_i,\ldots,x_d)-g_i(x_1,\ldots,x'_i,\ldots,x_d)) \geq 0.
    \end{multline*}
\end{itemize}
We emphasize that the notion of cyclical comonotonicity is not transitive in general, in contrast to the two other ones. For this reason, we avoid saying that several functions are cyclically comonotone to mean that they are pairwise cyclically comonotone.
\end{definition}

Another definition that will help formulate the problem concerns (marginal) probability distributions.

\begin{definition}[Graph-compatible distributions]\label{def:compatible}
Let $\G$ be a DAG. We say that a collection $(P_a)_{a \in \A}$ of $\calP(\R^d)$ is $\SCM^{\G}_{acy}-$compatible if there exists an $\M \in \SCM^{\G}_{acy}$ such that $(P_a)_{a \in \A}$ corresponds to the interventional marginals of $\M$. 
\end{definition}

In the upcoming analysis, we rely on the following properties of graph-compatible distributions.

\begin{lemma}[Factorization and existence]\label{lem:markov}
The two following propositions hold.
\begin{itemize}
    \item[$(i)$] Let $\G$ be a DAG. For every collection $(P_a)_{a \in \A}$ of $\calP(\R^d)$, if $(P_a)_{a \in \A}$ is $\SCM^{\G}_{acy}-$compatible, then for every $a \in \A$ and every $X_a \sim P_a$,
    \[
        \rmd \bbP(X_a=(x_1,\ldots,x_d)) = \prod^d_{i=1} \rmd \bbP(X_{a,i}=x_i | X_{a,\pa(i)} = x_{\pa(i)}).
    \]
    \item[$(ii)$] For every collection $(P_a)_{a \in \A}$ of $\calP(\R^d)$ such that $\A$ is countable, there exists a DAG $\G$ such that $(P_a)_{a \in \A}$ is $\SCM^{\G}_{acy}-$compatible.
\end{itemize}
\end{lemma}

\Cref{lem:markov} adapts several well-known results related to Markovian SCMs so that they fit our problem formulation. Notably, the factorization in case $(i)$ focuses only on the subset of variables $\{1,\ldots,d\}$ rather than on the whole $\{1,\ldots,d,\tta\}$, and addresses a collection of laws rather than a single endogenous distribution. We emphasize that case $(ii)$ of the above lemma states, in particular, that any collection of countably many marginals can be generated as the interventional marginals of some SCM. The countability condition is a simplifying technical assumption to discard measurability issues.

We now advance to the main results.

\subsubsection{General case}

The theorem below provides sufficient and necessary conditions on the SCM for its counterfactual maps to be some transport maps from Section~\ref{sec:transport}, in a general scenario. In the rest of the article, any equality between a counterfactual map $C_{a' \leftarrow a}$ and a transport map holds in a $P_a-$a.e sense while an equality between a subsolution map $g_a$ and a transport map holds in a $Q_{\ttx}-$a.e. sense. For clarity, we do not explicitly write almost-sure statements.

\begin{theorem}[Relations for general SCMs]\label{thm:main}
Let $\M \in \SCM$ be such that its interventional marginals $(P_a)_{a \in \A}$ have densities. Then,
\begin{itemize}
    \item[$(i)$] for every $a,a' \in \A$, $C_{a' \leftarrow a} = \CM(P_a,P_{a'})$ $\iff$ for every $a,a' \in \A$, $g_a$ and $g_{a'}$ are cyclically comonotone;
    \item[$(ii)$] for every $a,a' \in \A$, $C_{a' \leftarrow a} = \QP(P_a,P_{a'} ; p_0)$ $\iff$ there exists an injective map $\Phi \in \T(p_0,Q_{\ttx})$ such that, for every $a \in \A$, $g_a \circ \Phi$ is cyclically monotone.
\end{itemize}
\end{theorem}

In the above theorem, case $(i)$, cyclical comonotonicity appears as a natural generalization of the non-decreasing coupling introduced in Section \ref{sec:transport} for $d=1$. Cyclical comonotonicity ensures that the transports between marginal distributions preserve a global order structure analogous to monotone increasing maps, but extended to higher dimensions. 
In case $(ii)$, the condition concerns cyclical monotonicity after a common global transport given by $\Phi$. In other words, it is no longer just a matter of comparing the transports $g_a$ and $g_{a'}$ directly, but of recognizing that they become cyclically monotone once viewed through the common transformation $\Phi$. 
This highlights a conceptual difference: point $(i)$ emphasizes a property shared among the maps $(g_a)_{a \in \A}$, while point $(ii)$ introduces a universal reference frame (linked to $p_0$) that aligns all the maps so that they satisfy cyclical monotonicity.

\begin{remark}[Is there an SCM fitting my transport maps?]\label{rem:SCM_vs_transport}
    \Cref{thm:main} presumes the existence of an SCM, and then specifies sufficient and necessary conditions for its counterfactual maps to be CM or QP transports. The clarify the scope of this result, we take a different angle where the starting point is a collection of transport maps (meant to define counterfactuals) rather than an SCM.

    Given a family of countably many probability measures $(P_a)_{a \in \A}$, there always exists an SCM producing them (as interventional marginals) according to \cref{lem:markov}. However, given a family of transport maps $(T_{a' \leftarrow a})_{a,a' \in \A}$ between these marginals, there may not exist an SCM such that $C_{a' \leftarrow a} = T_{a' \leftarrow a}$ for every $a,a' \in \A$. Said differently, a collection of countably many marginals is always a collection of interventional marginals; a collection of couplings across these marginals is not always a collection of counterfactual couplings.
    
    In particular, according to \cref{prop:algebraic_ctf}, counterfactual maps (as induced by SCMs) meet all the properties from \cref{def:algebraic}. Moreover, according to \cref{prop:algebraic_classic}, a family of CM transport maps between preassigned marginals do not necessarily satisfy the path independence property from \cref{def:algebraic}. Therefore, given a collection of marginals, it may be impossible for the family of CM transport maps to be the counterfactual maps of some SCM. In other words, in many situations, one cannot define counterfactuals (in Pearl's sense) via optimal transport. Note that \cite{dance2025counterfactual} also highlights the algebraic limitations of optimal transport maps for counterfactual reasoning.

    In contrast, QP and KR transports always satisfies the algebraic properties of counterfactual maps. As such, they are more compatible to structural counterfactual reasoning. An upcoming result, \cref{thm:KR}, shows that KR transport maps are necessarily the counterfactual maps of some SCM. The case of QP transport maps is more complicated. While $(\QP(P_a,P_{a'};p_0))_{a,a' \in \A}$ does meet the algebraic constraints, whether one can systematically derive a valid causal mechanism $(f_i)_{i \in \I}$ (in the sense of \cref{def:scm}) from it is an open question.
\end{remark}

We also emphasize that \cref{thm:main} does not provide insight on the case where $C_{a' \leftarrow a} = \KR(P_a,P_{a'})$ for every $a,a' \in \A$. In particular, the SCM does not need to be acyclic if the counterfactual maps are triangular. Indeed, if the SCM is acyclic, then the subsolution maps and counterfactual maps are triangular according to \cref{prop:acyclic}. However, there exist acyclic SCMs with triangular (non-diagonal) counterfactual maps and non-triangular subsolution maps, as exemplified below.

\begin{example}[Cyclic SCM with triangular counterfactual maps]
Consider the cyclic SCM corresponding to the assignments:
\[
\begin{cases}
    A = U_{\tta}\\
    X_1 = U_1\\
    X_2 = X_1 X_3 + U_2\\
    X_3 = A X_2 + U_3\\
\end{cases}.
\]
This model was studied in \citep[Example 3.5]{bongers2021foundations}, as an example of cyclic SCM uniquely solvable with respect to every subset of variables. We propose to analyze its counterfactuals. The subsolution maps are given by,
\[
g_a(u_1,u_2,u_3) := \begin{pmatrix}
    u_1 \\ \frac{u_1 u_3 + u_2}{1 - a' u_1} \\ \frac{a' u_2 + u_3}{1 - a' u_1}
\end{pmatrix},
\]
and their inverses by,
\[
g^{-1}_a(x_1,x_2,x_3) := \begin{pmatrix}
    x_1 \\ x_2 - x_1 x_3 \\ x_3 - a x_2
\end{pmatrix}.
\]
Therefore, they are not triangular. Moreover, the counterfactual maps are expressed as
\[
C_{a' \leftarrow a}(x_1,x_2,x_3) := \begin{pmatrix}
    x_1 \\ x_2 \frac{1- a x_1}{1- a'x_1} \\ x_3 + (a'-a) \frac{x_2}{1-a'x_1}
\end{pmatrix},
\]
and as such are triangular and non-diagonal.
\end{example}

\subsubsection{Acyclic case}

In practice, most analysts rely on acyclic models, since cyclic models are not necessarily solvable and are hard to interpret. We addressed the cyclic case beforehand to gain in generality and to illustrate (below in \cref{sec:specific}) that some transport maps from \cref{sec:transport} can be attained by cyclic SCMs.

Let us now turn to the acyclic case. As mentioned earlier, the subsolution maps and the counterfactuals maps of acyclic SCMs share a triangular structure. Case $(i)$ from \cref{thm:main} can be extended in this situation, and the case where counterfactual maps coincide with KR maps can be specified.

\begin{theorem}[Relations for acyclic SCMs]\label{thm:main_acy}
Let $\M \in \SCM_{acy}$ be such that its interventional marginals $(P_a)_{a \in \A}$ have densities. Then,
\begin{itemize}
    \item[$(i)$] for every $a,a' \in \A$, $C_{a' \leftarrow a} = \CM(P_a,P_{a'})$ $\iff$ for every $a,a' \in \A$, $C_{a' \leftarrow a}$ is diagonal non-decreasing $\implies$ for every $a,a' \in \A$, $C_{a' \leftarrow a} = \KR(P_a,P_{a'})$;
    \item[$(ii)$] for every $a,a' \in \A$, $C_{a' \leftarrow a} = \KR(P_a,P_{a'})$ $\iff$ for every $a \in \A$, the functions $(g_a)_{a \in \A}$ are diagonally comonotone.
\end{itemize}
\end{theorem}

Regarding case $(i)$ above, we point out that $C_{a' \leftarrow a} = \KR(P_a,P_{a'})$ does not imply $C_{a' \leftarrow a} = \CM(P_a,P_{a'})$ in general when $\M \in \SCM_{acy}$. Thereafter, we carry out a more specific analysis, separately for each type of transports.

\subsection{Specific analysis}\label{sec:specific}

\Cref{thm:main} and \cref{thm:main_acy} translate conditions on counterfactual maps into conditions on solution maps. To clarify what this means in terms of causal models, we need to study how conditions on solution maps impact the SCM, notably its graph and its mechanisms. In what follows, we do not provide complete characterizations for each type of transports. Rather, we provide sufficient conditions on SCMs for their counterfactual maps to be classical transport maps.

\subsubsection{CM counterfactuals}

To date, there is no explicit characterization of the SCMs that produce CM counterfactual maps. According to \citep[Example 4]{delara2024transport}, the counterfactual maps of linear additive acyclic SCMs are simply translations, and hence are cyclically monotone. Actually, translations are also TM functions, meaning that, in this linear scenario, the counterfactual maps are also KR transports. One can readily extend this sufficient condition to the cyclic case. 

\begin{proposition}[Counterfactuals of partially linear SCMs]\label{prop:ctf_linear_CM_KR}
If $\M \in \SCM$ is a partially linear additive SCM with $(P_a)_{a \in \A}$ as interventional marginals, then $C_{a' \leftarrow a} = \CM(P_a,P_{a'}) = \KR(P_a,P_{a'})$.
\end{proposition}

While linear models are widely studied and applied in the causality literature, there are too stringent in common problems. In particular, they produce interventional marginals differing only by their means, and thereby cannot fit more complex interventional data. This notably raises the question of whether cyclically monotone counterfactual maps can be induced by other types of SCMs than simply linear ones. \citep[Example 5]{delara2024transport} provides a very specific positive example; a more general result is still lacking.

To clarify the analysis, we discuss necessary conditions on the SCM for its counterfactual maps to be CM. First of all, we emphasize that there are two collisions of distinct origins in the above result. The equality $\CM(P_a,P_{a'}) = \KR(P_a,P_{a'})$ comes from the fact that the marginals $(P_a)_{a \in \A}$ differ only by their means; the equality between $C_{a' \leftarrow a}$ and these transports comes from the fact that the SCM is partially linear additive. To illustrate the difference, one could consider an SCM entailing other counterfactual maps between the same marginals, which would imply $C_{a' \leftarrow a} \neq \CM(P_a,P_{a'}) = \KR(P_a,P_{a'})$. Interestingly, \cref{thm:main_acy} generalizes the collisions from \cref{prop:ctf_linear} to possibly nonlinear models \emph{in the acyclic case}, but the scope of this extension seems limited. According to \cref{prop:TM-CM_maps}, CM counterfactual maps must be diagonal. As such, \emph{in the acyclic case, only SCMs producing diagonal counterfactual maps entail CM counterfactuals}. This remark combined with the fact that CM maps do not always meet the algebraic properties of counterfactual maps (\cref{rem:SCM_vs_transport}) upholds the idea that CM transports are not particularly tailored to counterfactual reasoning in the sense of Pearl. Whether this inconsistency with Pearl's causal framework is a limitation remains a debatable question; this is however a critical point to keep in mind, considering that such transports have become more common to define counterfactual maps \citep{charpentier2023optimal, torous2024optimal, delara2024transport, balakrishnan2025conservative}.

\subsubsection{QP counterfactuals}

Item $(ii)$ is arguably the most convoluted condition of \cref{thm:main}, as it involves an injective transport map $\Phi$. Critically, $\Phi$ does not need to correspond to an exogenous reparameterization of $\M$ (as in \cref{prop:reparameterization}) and therefore is not constrained to be diagonal. The fact that $\Phi$ can have almost any form makes it hard to identify the SCMs producing quantile-preserving counterfactuals.

However, interestingly, if $\Phi$ is diagonal, then the graph of the SCM has a very peculiar shape, as explained below.

\begin{proposition}[A class of QP linear SCMs]\label{prop:linear_QP}
Let $\M \in \SCM$ be a linear additive noise model such that its interventional marginals $(P_a)_{a \in \A}$ have densities. If $\M$ is such that,
\begin{itemize}
    \item[$(i)$] for every $a \in \A$, $g_a = \CM(p_0,P_a) \circ \Phi$, where $\Phi \in \T(Q_{\ttx},p_0)$ is injective and diagonal,
    \item[$(ii)$] for every $a \in \A$, $g_a$, $\CM(p_0,P_a)$ and $\Phi$ are differentiable,
\end{itemize}
then $j \in \pa(i) \iff i \in \pa(j)$.
\end{proposition}

Note that the \emph{everywhere} differentiability assumption of all involved functions handles the fact that differentiability is not preserved under null sets pertubations, as mentioned in the proof of \cref{prop:TM-CM_maps}.

Interestingly, the above result holds regardless of the reference measure defining the quantiles. Under the assumptions of \cref{prop:linear_QP}, the graph of $\M$ has either no arrows or is cyclic with very specific cycles: every arrow involving a variable in $[d]$ has a converse counterpart, producing short loops. We point out that such SCMs exist.

\begin{example}[A linear quantile-preserving SCM]
Choose $p_0 = \square$, and let $\M \in \SCM$ be the SCM corresponding to the assignments
\begin{align*}
    &A = U_{\tta},\\
    &X_1 = A - \frac{1}{3} X_1 + \frac{2}{3} X_2 + U_1,\\
    &X_2 = A + \frac{2}{3} X_1 - \frac{1}{3} X_2 + U_2,
\end{align*}
where $(U_1,U_2) \sim \square$. Note that $\M$ is cyclic and uniquely solvable with respect to $\{1,2\}$, such that
\[
    g_a(u_1,u_2) = \frac{3a}{2} \begin{pmatrix}
        1 \\ 1
    \end{pmatrix} + \begin{pmatrix}
        1 & 1/2 \\ 1/2 & 1
    \end{pmatrix} \begin{pmatrix}
        u_1 \\ u_2
    \end{pmatrix}.
\]
One can check that $g_a$ is CM. Therefore, since $p_0 = \square = Q_{\ttx}$, $g_a = \CM(\square,P_a)$. The SCM meets the assumptions of \cref{prop:linear_QP} with $\Phi = \Id$, so that $C_{a' \leftarrow a} = \QP(P_a,P_{a'} ; \square)$, and the graph is indeed cyclic.

Note also that $C_{a' \leftarrow a} = \CM(P_a,P_{a'}) = \KR(P_a,P_{a'})$ according to \cref{prop:ctf_linear}. Consequently, in this specific example, $C_{a' \leftarrow a} = \CM(P_a,P_{a'}) = \KR(P_a,P_{a'}) = \QP(P_a,P_{a'} ; \square)$.
\end{example}

In the proof of \cref{prop:linear_QP}, the linearity assumption serves to explicitly trace back $\Jac(g_a)$ to the matrix $M^{\ttx}$ describing the SCM. We can extend the key graphical conclusions of the proposition to nonlinear SCMs by supposing that $\frac{\partial g_{a,i}}{\partial u_j} \not\equiv 0 \iff j \in \an(i)$ for all $a \in \A$. We emphasize that this is a weak assumption. In the acyclic case, $g_a$ has a closed form, where each $g_{a,i}$ is a function of only $u_{\an(i)}$ (recall \cref{prop:acyclic}). As such, this assumption demands $u_j$ for every $j \in \an(i)$ to be a true argument of $g_{a,i}$, which simply requires each $f_i$ to be a nonconstant function of every of its inputs. Whether a similar dependence relation between $g_{a,i}$ and $\an(i)$ holds in the cyclic case remains an open question. The proposition below generalizes \cref{prop:linear_QP}. 

\begin{proposition}[A class of QP SCMs]\label{prop:QP_SCM}
Let $\M \in \SCM$ with $(P_a)_{a \in \A}$ as interventional marginals. Suppose that $g_a$ is differentiable for every $a \in \A$ and is such that $(\Jac(g_a)_{i,j} \neq 0 \iff j \in \an(i))$. If $\M$ is such that,
\begin{itemize}
    \item[$(i)$] for every $a \in \A$, $g_a = \CM(p_0,P_a) \circ \Phi$, where $\Phi \in \T(Q_{\ttx},p_0)$ is injective and diagonal,
    \item[$(ii)$] for every $a \in \A$, $g_a$, $\CM(p_0,P_a)$ and $\Phi$ are differentiable,
\end{itemize}
then $j \in \an(i) \iff i \in \an(j)$.
\end{proposition}

As before, this implies that the graph of $\M$ is either cyclic (with peculiar cycles) or edgeless. \Cref{prop:linear_QP,prop:QP_SCM} call for further investigation. Notably, it would be interesting to see what happens when $\Phi$ is not restricted to be diagonal.

\subsubsection{KR counterfactuals}

We finally address KR counterfactuals. Before discussing monotonicity, we recall that the counterfactual maps of ayclic SCMs are triangular. In general, however, the counterfactual maps of acyclic SCMs do not need to be triangular \emph{monotone}. The proposition below extends a well-known sufficient condition under which an SCM yields KR counterfactual maps.

\begin{proposition}[SCMs with KR counterfactuals]\label{prop:KR_ctf}
Let $\M \in \SCM_{acy}$ with $(P_a)_{a \in \A}$ as interventional marginals. If, for every $i \in [d]$, the functions in $\{f_i(a_{\pa(i)},x_{\pa(i)},\cdot)\}_{(x,a) \in \X \times \A}$ are comonotone, then $C_{a' \leftarrow a} = \KR(P_a,P_{a'})$ for every $a,a' \in \A$. 
\end{proposition}

We point out that \cref{prop:ctf_linear} is a specific case of the above result, as the mechanisms of partially linear additive models are increasing in their exogenous variables. In this scenario, the counterfactual maps are diagonal increasing, which is more specific than being TM.

\Cref{prop:KR_ctf} furnishes a general sufficient criterion on the SCM for the counterfactual maps to be KR. Conversely, one can always find a modification of an SCM that preserves the interventional marginals and renders the counterfactual maps TM, as formalized below.

\begin{proposition}[Existence of KR counterfactuals]\label{prop:KR_representation}
Let $(P_a)_{a \in \A}$ be a family elements of $\calP(\R^d)$ with densities and $\G$ be a DAG. If $(P_a)_{a \in \A}$ is $\SCM^{\G}_{acy}-$compatible, then there exists $\M \in \SCM^{\G}_{acy}$ with $(P_a)_{a \in \A}$ as interventional marginals and such that $C_{a' \leftarrow a} = \KR(P_a,P_{a'})$.   
\end{proposition}

Recall that, according to \cref{lem:markov}, every collection $(P_a)_{a \in \A}$ with countable $\A$ can be generated by some acyclic SCM. As such, \cref{prop:KR_representation} signifies that $(\KR(P_a,P_{a'}))_{a,a' \in \A}$ are always valid counterfactual maps across these marginals, in the sense that there exists an SCM producing them. As mentioned earlier, this is critically not the case for $(\CM(P_a,P_{a'}))_{a,a' \in \A}$ and the question is open for $(\QP(P_a,P_{a'};p_0))_{a,a' \in \A}$ (\cref{rem:SCM_vs_transport}).

The above results underline natural connections between KR transport maps and counterfactual maps induced by acyclic SCMs. This explains why the causality literature has often considered KR counterfactuals, either indirectly by specifying the causal model or directly by specifying the counterfactual maps. More concretely, some articles define the latent SCM so that $g_a$ is TM for every $a \in \A$, which renders $C_{a' \leftarrow a}$ TM for every $a,a' \in \A$ by stability under composition \citep{ma2006quantile,plevcko2020fair,javaloy2024causal}; other define the couterfactual maps as TM transport maps \citep{machado2025sequential}.

\subsection{Take home message: on using canonical transports to define counterfactuals}

Some recent papers have proposed to circumvent the full specification of a causal model when reasoning counterfactually by directly placing transport maps between interventional marginals \citep{black2020fliptest, charpentier2023optimal, torous2024optimal, delara2024transport, machado2025sequential, dance2025counterfactual}. Critically, imposing the counterfactual maps to meet certain properties has implications on the latent causal models, and thereby on the causal assumptions. As such, we argue that analysts must verify that their choices of transport maps do not clash with the basic causal assumptions they have in mind.

Say, for instance, that an analyst does not have a precise idea of the causal graph governing their problem, and simply postulates that the graph is a DAG. Then, they define the counterfactual maps as the CM maps between the interventional marginals (assuming they are identifiable). This raises several issues. First, there is no guarantee that a latent SCM corresponding to such maps exist, as explained in \cref{rem:SCM_vs_transport}. Second, according to \cref{thm:main_acy}, the counterfactual maps need to be diagonal increasing, which considerably reduces the class of compatible SCMs. This is why choosing CM counterfactuals without further knowledge of the data-generating process does not seem very reliable. Selecting quantile-preserving counterfactuals raises a similar issue: in some cases (see \cref{prop:QP_SCM}) the causal graph must be cyclic (even with very particular loops) or without edges, which contradicts standard graphical conceptions. As soon as compatibility to the \emph{acyclic}-SCM framework matters, the only systematically valid monotone transport is KR (\cref{thm:KR}). This feature, combined with the fact that KR maps are limits of optimal transport maps \citep{carlier2010knothe} led \cite{backhoff2017causal} to consider KR maps as causal versions of CM maps. Nonetheless, applying TM maps in a counterfactual context requires to suppose a causal ordering. We refer to \cite{dance2025counterfactual, delara2025canonical} for more general constructions of counterfactual couplings that are coherent by design with acyclic SCMs.

All in all, we caution against arbitrary choices and intuitions of transport maps to define counterfactuals. Typically, statisticians find it natural to align distributions via their quantiles. Such an intuition is perfectly valid from a observational perspective but not necessarily from a causal perspective, as previously explained.

\section{Further discussions}\label{sec:further}

To conclude this article, we further discuss counterfactual couplings and related questions.

\subsection{On the scope of our causal framework}

In the causality literature, counterfactual problems can be formalized from a different angle than ours, notably with other types of causal models. Hereafter, we discuss these aspects in details to clarify the scope of the analysis we carried out in \cref{sec:counterfactual} and \cref{sec:meet}.

\subsubsection{About counterfactual couplings}

We presented counterfactual couplings and counterfactual maps as derivatives of the distributions induced by $(X_a,X_{a'})$ for every $a,a' \in \A$. Note that, while such pairs do describe two parallel worlds, they do not distinguish the actual one from the counterfactual one. This is why some analysts prefer the coupling $\law{(X,X_{a'})|A=a}$, which places probabilities on all contrary-to-fact statements of the form \say{Had $A$ been equal to $a'$ instead of $a$, then $X$ would have been equal to $x'$ instead of $x$}. It notably underpins definitions of counterfactual fairness \citep{kusner2017counterfactual}.

We emphasize that our formalism captures this perspective. Crucially, a coupling like $\law{(X,X_{a'})|A=a}$, for some $a,a' \in \A$, is directly derived from $\law{X_a,X_{a'}}$ and corresponds to the same counterfactual map. Indeed, under \cref{hyp:scms},
\begin{align*}
    \law{(X,X_{a'}) | A=a} &= \law{(X,g_{a'}(U_{\ttx})) | A=a} \ \text{by definition of $g_{a'}$,}\\
    &= \law{(X, \left(g_{a'} \circ g^{-1}_a\right)(X_a) | A=a} \ \text{by inversion,}\\
    &= \law{(X, \left(g_{a'} \circ g^{-1}_a\right)(X) | A=a} \ \text{by consistency,}\\
    &= \left(\Id, \left(g_{a'} \circ g^{-1}_a\right)\right) \sharp \law{X | A=a} \ \text{by transfer.}
\end{align*}
Thereby, $\law{(X,X_{a'})|A=a}$ is a deterministic coupling induced by the counterfactual map $C_{a' \leftarrow a}$. We point out that the underlying push-forward constraint is here $$\law{X_{a'}|A=a} = {C_{a' \leftarrow a}} \sharp \law{X|A=a},$$ which differs from $P_{a'} = {C_{a' \leftarrow a}} \sharp P_a$. The validity of applying this same function between distinct marginals comes from the fact that the supports of $\law{X_{a'}|A=a}$ and $\law{X|A=a}$ are, respectively, included in the ones of $P_{a'}$ and $P_{a}$.

Note that deterministic counterfactual couplings correspond to deterministic counterfactual statements, in the sense that \say{Had $A$ been equal to $a'$ instead of $a$, then $X$ would have been equal to $x'$ instead of $x$} has probability one if $x' = C_{a' \leftarrow a}(x)$ and zero otherwise.

\subsubsection{About the potential-outcome framework}

Another classical approach to causal modeling is the potential-outcome framework of \cite{neyman1923applications} and \cite{rubin1974estimating}. Many articles from this field define every postintervention variables as generated by a deterministic function (sometimes called a \emph{production function}) of a latent variable \citep{torous2024optimal}. This follows the same principle as the equations $X_a = g_a(U_{\ttx})$ (for every $a \in \A$) that we obtained by postulating a latent SCM. As such, and as expected by the causality community \citep{pearl2009causality,ibeling2023comparing,delara2025clarification}, the two frameworks can be made equivalent. The main difference lies in whether the production functions are primitives (potential-outcome framework) or derivatives (structural framework) of a causal model. For instance, \cref{prop:submap} would be an assumption rather than a proposition in the potential-outcome framework.\footnote{We refer to \cite{pearl2010brief} for further details on the primitive-v.s.-derivative question.} More generally, all our assumptions and theorems that are framed as properties of $(g_a)_{a \in \A}$ only can be translated into the potential-outcome vocabulary (like \cref{thm:main}). The analysis from \cref{sec:specific} specifically depends on Pearl's causal framework, as it determines SCMs that produce solution maps satisfying the considered properties.

In this work, we rather rely on SCMs because they furnish a more intelligible formalization of causal assumptions (via graphs and cause-effect equations). Notably, this enables us to provide interpretable implications of counterfactual choices.

\subsection{Related works}

In this subsection, we clarify the scope of our work by briefly reviewing related articles.

\cite{delara2024transport} introduced a framework allowing the use of any transport maps as counterfactual maps and then focused on the choice of optimal transport maps for the squared Euclidean cost, which coincide with CM maps. Their inspiration comes from \cite{black2020fliptest}, who seminally employed optimal transport maps for counterfactual reasoning (although for slightly distinct cost functions). Other articles proposed to use CM maps as counterfactual maps, arguing that they generalize monotonicity to higher dimensions \citep{charpentier2023optimal,torous2024optimal}. \cite{machado2025sequential} noted the importance of respecting the topological order on the variables (when a correct DAG is accessible) and thereby preferred KR maps. Additionally, other work relied (sometimes implicitly) on KR maps as counterfactual maps \citep{hoderlein2016testing,plevcko2020fair,javaloy2024causal}, by defining the structural assignments via \emph{conditional} (univariate) quantile functions. As far as we know, the idea of relying on \emph{multivariate} QP maps for counterfactual reasoning is original. We believe it was a natural option to explore in the context of extending the univariate monotone rearrangement. All in all, these contributions put forward two ideas: circumventing the full specification of a causal model by considering a mass-transportation viewpoint of counterfactual reasoning; choosing relevant transport maps via optimality or monotonicity criteria. This growing literature underpins the main motivation of our work, that is, clarifying the causal assumptions that analysts make when choosing a specific sort of transport maps for counterfactual maps.

In this sense, our article complements the work of \cite{dance2025counterfactual}. They introduce a framework for constructing transport maps that can always be viewed as the counterfactual maps of some SCM by integrating algebraic properties in their design. In doing so, they develop robust estimation procedures of counterfactual couplings, but do not address the arbitrariness of the coupling choice. In contrast, our work focuses on theoretical aspects, and relates counterfactual maps to classical transport maps (which do not necessarily stem from SCMs) to highlight the distinction between causal features and statistical intuitions. Additionally, our framework allows the graph to be cyclic and the manipulated variable (namely, $\tta$) to have \emph{any} position in the graph.

We also point out that several articles employ transport frameworks in causality but address radically distinct problems. For instance, \cite{akbari2023causal} uses KR maps as a tool for causal discovery---not for counterfactual reasoning. Moreover, a series of papers applies optimal transport for causal discovery or interventional inference \citep{tu2022optimal,cheridito2025optimal,lin2025tightening}. To better classify the literature in general, we recommend people to ask two questions when reading papers bridging optimal transport and causality: what is the optimal-transport solution at play (the distance or the transport plan) and what is the causality layer of interest (interventional or counterfactual).

\subsection{Open questions}

The analysis from \cref{sec:meet} deserves further research. In particular, there is currently no precise characterization of SCMs inducing CM counterfactuals. Similarly, QP SCMs have not been fully identified. As aforementioned, it would be interesting to provide results like \cref{prop:linear_QP} and \cref{prop:QP_SCM} for a general transformation $\Phi$ (not necessarily diagonal).

An aspect which is not discussed in this paper is the computational cost of computing classical transports (for instance to define counterfactuals). Because they all rely on solving optimal transport problems (for different distributions and cost), $\CM$, $\QP$ and $\KR$ maps all raise computational issues. A celebrated circumvention of \cite{cuturi2013sinkhorn} for faster computation is to relax the transport condition and consider, for instance for $\KR$, another cost which is the interpolation between the $\|\cdot\|^2_{\eps}$ transport term and an entropic regularization term, for small $\eps>0$. Finding the optimal coupling for this alternative cost is computationally tractable. The main question, however, is whether such intermediate constructions admit a clear causal interpretation. We also point out that entropic regularization blends the transport maps, making it a random coupling. One can either use barycentric-projection techniques \citep{pooladian2021entropic} to recover a deterministic coupling (that is, a transport map) or work with the obtained random coupling.

This opens another direction, which would be to explore non-deterministic couplings for counterfactuals -- that is, joint distributions between $P_a$ and $P_{a'}$ which are not simply of the form $(\Id,T){\sharp}P_a$. Considering such couplings would allow to gain in generality, at the price of thornier identifiability questions. Note that the frameworks from \citep{delara2024transport} and \citep{delara2025canonical} allow stochastic couplings as counterfactual couplings.

\section{Conclusion}

One must always keep in mind that there exist various ways of matching two probability measures. The scientific literature tends to consider that adequate transport maps should be monotone or optimal. However, these criteria are rather imprecise in dimensions higher than one. We proposed a comparative study of three way of matching measures corresponding to these criteria, which highlights fundamental differences (\cref{tab:transports_comparison}) but also cases of coincidence (\cref{sec:collision}). We hope this will provide better guidance to analysts when selecting a \say{good} transport map, in particular make them consider alternatives to optimal transport maps. To push the analysis, we compared these three types of transport in the context of counterfactual reasoning. This notably underlines significant limitations on the use of CM and QP transports, in contrast to KR transports. We expect this will clarify the scope and implications of recent frameworks that relied on CM maps or other transport maps to compute counterfactual coupling. Moreover, this causal detour allows to better understand fundamental features of classical transports in statistical applications.

\subsection*{Acknowledgments}
This work was partially funded by the project \textsc{CAUSALI-T-AI} (ANR-23-PEIA-0007) of the
French National Research Agency. 

\bibliographystyle{abbrvnat}
\bibliography{references}

\begin{thebibliography}{80}
\providecommand{\natexlab}[1]{#1}
\providecommand{\url}[1]{\texttt{#1}}
\expandafter\ifx\csname urlstyle\endcsname\relax
  \providecommand{\doi}[1]{doi: #1}\else
  \providecommand{\doi}{doi: \begingroup \urlstyle{rm}\Url}\fi

\bibitem[Akbari et~al.(2023)Akbari, Ganassali, and Kiyavash]{akbari2023causal}
S.~Akbari, L.~Ganassali, and N.~Kiyavash.
\newblock Causal discovery via monotone triangular transport maps.
\newblock In \emph{NeurIPS 2023 Workshop Optimal Transport and Machine
  Learning}, 2023.

\bibitem[Athey and Imbens(2006)]{athey_CIC_06}
S.~Athey and G.~W. Imbens.
\newblock Identification and {Inference} in {Nonlinear}
  {Difference}-in-{Differences} {Models}.
\newblock \emph{Econometrica}, 74\penalty0 (2):\penalty0 431--497, Mar. 2006.
\newblock ISSN 0012-9682, 1468-0262.
\newblock \doi{10.1111/j.1468-0262.2006.00668.x}.
\newblock URL \url{http://doi.wiley.com/10.1111/j.1468-0262.2006.00668.x}.

\bibitem[Backhoff et~al.(2017)Backhoff, Beiglbock, Lin, and
  Zalashko]{backhoff2017causal}
J.~Backhoff, M.~Beiglbock, Y.~Lin, and A.~Zalashko.
\newblock Causal transport in discrete time and applications.
\newblock \emph{SIAM Journal on Optimization}, 27\penalty0 (4):\penalty0
  2528--2562, 2017.

\bibitem[Balakrishnan et~al.(2025)Balakrishnan, Kennedy, and
  Wasserman]{balakrishnan2025conservative}
S.~Balakrishnan, E.~Kennedy, and L.~Wasserman.
\newblock Conservative inference for counterfactuals.
\newblock \emph{Journal of Causal Inference}, 13\penalty0 (1):\penalty0
  20230071, 2025.

\bibitem[Baptista et~al.(2024)Baptista, Marzouk, and
  Zahm]{baptista2024representation}
R.~Baptista, Y.~Marzouk, and O.~Zahm.
\newblock On the representation and learning of monotone triangular transport
  maps.
\newblock \emph{Foundations of Computational Mathematics}, 24\penalty0
  (6):\penalty0 2063--2108, 2024.

\bibitem[Bartl et~al.(2024)Bartl, Beiglb{\"o}ck, and
  Pammer]{bartl2024wasserstein}
D.~Bartl, M.~Beiglb{\"o}ck, and G.~Pammer.
\newblock The {W}asserstein space of stochastic processes.
\newblock \emph{Journal of the European Mathematical Society}, 2024.

\bibitem[Beirlant et~al.(2020)Beirlant, Buitendag, {del Barrio}, Hallin, and
  Kamper]{beirlant2020center}
J.~Beirlant, S.~Buitendag, E.~{del Barrio}, M.~Hallin, and F.~Kamper.
\newblock Center-outward quantiles and the measurement of multivariate risk.
\newblock \emph{Insurance: Mathematics and Economics}, 95:\penalty0 79--100,
  2020.

\bibitem[Black et~al.(2020)Black, Yeom, and Fredrikson]{black2020fliptest}
E.~Black, S.~Yeom, and M.~Fredrikson.
\newblock Fliptest: fairness testing via optimal transport.
\newblock In \emph{Proceedings of the 2020 conference on fairness,
  accountability, and transparency}, pages 111--121, 2020.

\bibitem[Bogachev et~al.(2005)Bogachev, Kolesnikov, and
  Medvedev]{bogachev2005triangular}
V.~I. Bogachev, A.~V. Kolesnikov, and K.~V. Medvedev.
\newblock Triangular transformations of measures.
\newblock \emph{Sbornik: Mathematics}, 196\penalty0 (3):\penalty0 309, 2005.

\bibitem[Boissard et~al.(2015)Boissard, Gouic, and
  Loubes]{boissard2015distribution}
E.~Boissard, T.~L. Gouic, and J.-M. Loubes.
\newblock {Distribution’s template estimate with {W}asserstein metrics}.
\newblock \emph{Bernoulli}, 21\penalty0 (2):\penalty0 740 -- 759, 2015.
\newblock \doi{10.3150/13-BEJ585}.
\newblock URL \url{https://doi.org/10.3150/13-BEJ585}.

\bibitem[Bongers et~al.(2021)Bongers, Forr{\'e}, Peters, and
  Mooij]{bongers2021foundations}
S.~Bongers, P.~Forr{\'e}, J.~Peters, and J.~M. Mooij.
\newblock Foundations of structural causal models with cycles and latent
  variables.
\newblock \emph{The Annals of Statistics}, 49\penalty0 (5):\penalty0
  2885--2915, 2021.

\bibitem[Brenier(1991)]{brenier1991polar}
Y.~Brenier.
\newblock Polar factorization and monotone rearrangement of vector-valued
  functions.
\newblock \emph{Communications on pure and applied mathematics}, 44\penalty0
  (4):\penalty0 375--417, 1991.

\bibitem[Carlier et~al.(2010)Carlier, Galichon, and
  Santambrogio]{carlier2010knothe}
G.~Carlier, A.~Galichon, and F.~Santambrogio.
\newblock From knothe's transport to brenier's map and a continuation method
  for optimal transport.
\newblock \emph{SIAM Journal on Mathematical Analysis}, 41\penalty0
  (6):\penalty0 2554--2576, 2010.

\bibitem[Carlier et~al.(2016)Carlier, Chernozhukov, and
  Galichon]{carlier2016vector}
G.~Carlier, V.~Chernozhukov, and A.~Galichon.
\newblock Vector quantile regression: an optimal transport approach.
\newblock \emph{The Annals of Statistics}, 44\penalty0 (3):\penalty0
  1165--1192, 2016.

\bibitem[Chang et~al.(2022)Chang, Shi, Tuan, and Wang]{chang2022unified}
W.~Chang, Y.~Shi, H.~Tuan, and J.~Wang.
\newblock Unified optimal transport framework for universal domain adaptation.
\newblock \emph{Advances in Neural Information Processing Systems},
  35:\penalty0 29512--29524, 2022.

\bibitem[Charpentier et~al.(2023)Charpentier, Flachaire, and
  Gallic]{charpentier2023optimal}
A.~Charpentier, E.~Flachaire, and E.~Gallic.
\newblock Optimal transport for counterfactual estimation: A method for causal
  inference.
\newblock In \emph{Optimal Transport Statistics for Economics and Related
  Topics}, pages 45--89. Springer, 2023.

\bibitem[Chen and Du(2025)]{chen2025exogenous}
Y.~Chen and D.~Du.
\newblock Exogenous isomorphism for counterfactual identifiability.
\newblock \emph{arXiv preprint arXiv:2505.02212}, 2025.

\bibitem[Cheridito and Eckstein(2025)]{cheridito2025optimal}
P.~Cheridito and S.~Eckstein.
\newblock Optimal transport and {W}asserstein distances for causal models.
\newblock \emph{Bernoulli}, 31\penalty0 (2):\penalty0 1351--1376, 2025.

\bibitem[Chernozhukov et~al.(2015)Chernozhukov, Galichon, Hallin, and
  Henry]{chernozhukov2015}
V.~Chernozhukov, A.~Galichon, M.~Hallin, and M.~Henry.
\newblock Monge-kantorovich depth, quantiles, ranks, and signs, 2015.
\newblock URL \url{https://arxiv.org/abs/1412.8434}.

\bibitem[Courty et~al.(2016)Courty, Flamary, Tuia, and
  Rakotomamonjy]{courty2016optimal}
N.~Courty, R.~Flamary, D.~Tuia, and A.~Rakotomamonjy.
\newblock Optimal transport for domain adaptation.
\newblock \emph{IEEE transactions on pattern analysis and machine
  intelligence}, 39\penalty0 (9):\penalty0 1853--1865, 2016.

\bibitem[Cuesta and Matr{\'a}n(1989)]{cuesta1989notes}
J.~A. Cuesta and C.~Matr{\'a}n.
\newblock Notes on the {W}asserstein metric in hilbert spaces.
\newblock \emph{The Annals of Probability}, pages 1264--1276, 1989.

\bibitem[Cuturi(2013)]{cuturi2013sinkhorn}
M.~Cuturi.
\newblock Sinkhorn distances: Lightspeed computation of optimal transport.
\newblock \emph{Advances in neural information processing systems}, 26, 2013.

\bibitem[Dance and Bloem-Reddy(2025)]{dance2025counterfactual}
H.~Dance and B.~Bloem-Reddy.
\newblock Counterfactual cocycles: A framework for robust and coherent
  counterfactual transports.
\newblock \emph{arXiv preprint arXiv:2405.13844}, 2025.

\bibitem[Davies and Soundy(2009)]{davies2009genetics}
G.~E. Davies and T.~J. Soundy.
\newblock The genetics of smoking and nicotine addiction.
\newblock \emph{South Dakota Medicine}, 2009.

\bibitem[Dawid(2000)]{dawid2000causal}
A.~P. Dawid.
\newblock Causal inference without counterfactuals.
\newblock \emph{Journal of the American statistical Association}, 95\penalty0
  (450):\penalty0 407--424, 2000.

\bibitem[De~Lara(2025{\natexlab{a}})]{delara2025canonical}
L.~De~Lara.
\newblock Canonical representations of markovian structural causal models: A
  framework for counterfactual reasoning.
\newblock \emph{arXiv preprint arXiv:2507.16370}, 2025{\natexlab{a}}.

\bibitem[De~Lara(2025{\natexlab{b}})]{delara2025clarification}
L.~De~Lara.
\newblock A clarification on the links between potential outcomes and
  do-interventions.
\newblock \emph{Journal of Causal Inference}, 13\penalty0 (1):\penalty0
  20240033, 2025{\natexlab{b}}.
\newblock \doi{doi:10.1515/jci-2024-0033}.
\newblock URL \url{https://doi.org/10.1515/jci-2024-0033}.

\bibitem[De~Lara et~al.(2021)De~Lara, Gonz{\'a}lez-Sanz, and
  Loubes]{delara2021consistent}
L.~De~Lara, A.~Gonz{\'a}lez-Sanz, and J.-M. Loubes.
\newblock A consistent extension of discrete optimal transport maps for machine
  learning applications.
\newblock \emph{arXiv preprint arXiv:2102.08644}, 2021.

\bibitem[De~Lara et~al.(2024)De~Lara, Gonz{\'a}lez-Sanz, Asher, Risser, and
  Loubes]{delara2024transport}
L.~De~Lara, A.~Gonz{\'a}lez-Sanz, N.~Asher, L.~Risser, and J.-M. Loubes.
\newblock Transport-based counterfactual models.
\newblock \emph{Journal of Machine Learning Research}, 25\penalty0
  (136):\penalty0 1--59, 2024.

\bibitem[De~Lara et~al.(2025)De~Lara, Deronzier, Gonz\'{a}lez-Sanz, and
  Foy]{delara2025nonconvexity}
L.~De~Lara, M.~Deronzier, A.~Gonz\'{a}lez-Sanz, and V.~Foy.
\newblock On the nonconvexity of push-forward constraints and its consequences
  in machine learning.
\newblock \emph{SIAM Journal on Mathematics of Data Science}, 7\penalty0
  (2):\penalty0 597--620, 2025.
\newblock \doi{10.1137/24M1645036}.
\newblock URL \url{https://doi.org/10.1137/24M1645036}.

\bibitem[Flamary et~al.(2021)Flamary, Courty, Gramfort, Alaya, Boisbunon,
  Chambon, Chapel, Corenflos, Fatras, Fournier, Gautheron, Gayraud, Janati,
  Rakotomamonjy, Redko, Rolet, Schutz, Seguy, Sutherland, Tavenard, Tong, and
  Vayer]{flamary2021pot}
R.~Flamary, N.~Courty, A.~Gramfort, M.~Z. Alaya, A.~Boisbunon, S.~Chambon,
  L.~Chapel, A.~Corenflos, K.~Fatras, N.~Fournier, L.~Gautheron, N.~T. Gayraud,
  H.~Janati, A.~Rakotomamonjy, I.~Redko, A.~Rolet, A.~Schutz, V.~Seguy, D.~J.
  Sutherland, R.~Tavenard, A.~Tong, and T.~Vayer.
\newblock Pot: Python optimal transport.
\newblock \emph{Journal of Machine Learning Research}, 22\penalty0
  (78):\penalty0 1--8, 2021.
\newblock URL \url{http://jmlr.org/papers/v22/20-451.html}.

\bibitem[Gayraud et~al.(2017)Gayraud, Rakotomamonjy, and
  Clerc]{gayraud2017optimal}
N.~T. Gayraud, A.~Rakotomamonjy, and M.~Clerc.
\newblock Optimal transport applied to transfer learning for p300 detection.
\newblock In \emph{BCI 2017-7th Graz Brain-Computer Interface Conference},
  page~6, 2017.

\bibitem[Gonz{\'a}lez-Sanz et~al.(2022)Gonz{\'a}lez-Sanz, De~Lara, B{\'e}thune,
  and Loubes]{gonzalez2022gan}
A.~Gonz{\'a}lez-Sanz, L.~De~Lara, L.~B{\'e}thune, and J.-M. Loubes.
\newblock Gan estimation of lipschitz optimal transport maps.
\newblock \emph{arXiv preprint arXiv:2202.07965}, 2022.

\bibitem[Hallin et~al.(2021)Hallin, del Barrio, Cuesta-Albertos, and
  Matr{\'a}n]{hallin2021distribution}
M.~Hallin, E.~del Barrio, J.~Cuesta-Albertos, and C.~Matr{\'a}n.
\newblock {Distribution and quantile functions, ranks and signs in dimension d:
  A measure transportation approach}.
\newblock \emph{The Annals of Statistics}, 49\penalty0 (2):\penalty0 1139 --
  1165, 2021.
\newblock \doi{10.1214/20-AOS1996}.
\newblock URL \url{https://doi.org/10.1214/20-AOS1996}.

\bibitem[Hoderlein et~al.(2016)Hoderlein, Su, White, and
  Yang]{hoderlein2016testing}
S.~Hoderlein, L.~Su, H.~White, and T.~T. Yang.
\newblock Testing for monotonicity in unobservables under unconfoundedness.
\newblock \emph{Journal of Econometrics}, 193\penalty0 (1):\penalty0 183--202,
  2016.

\bibitem[Huang et~al.(2021)Huang, Chen, Tsirigotis, and
  Courville]{huang2021convex}
C.-W. Huang, R.~T.~Q. Chen, C.~Tsirigotis, and A.~Courville.
\newblock Convex potential flows: Universal probability distributions with
  optimal transport and convex optimization.
\newblock In \emph{International Conference on Learning Representations}, 2021.
\newblock URL \url{https://openreview.net/forum?id=te7PVH1sPxJ}.

\bibitem[Ibeling and Icard(2023)]{ibeling2023comparing}
D.~Ibeling and T.~Icard.
\newblock Comparing causal frameworks: Potential outcomes, structural models,
  graphs, and abstractions.
\newblock \emph{Advances in Neural Information Processing Systems}, 37, 2023.

\bibitem[Javaloy et~al.(2024)Javaloy, S{\'a}nchez-Mart{\'\i}n, and
  Valera]{javaloy2024causal}
A.~Javaloy, P.~S{\'a}nchez-Mart{\'\i}n, and I.~Valera.
\newblock Causal normalizing flows: from theory to practice.
\newblock \emph{Advances in Neural Information Processing Systems}, 36, 2024.

\bibitem[Kallenberg(1997)]{kallenberg1997foundations}
O.~Kallenberg.
\newblock \emph{Foundations of modern probability}.
\newblock Springer, 1997.

\bibitem[Karimi et~al.(2021)Karimi, Sch{\"o}lkopf, and
  Valera]{karimi2021algorithmic}
A.-H. Karimi, B.~Sch{\"o}lkopf, and I.~Valera.
\newblock Algorithmic recourse: from counterfactual explanations to
  interventions.
\newblock In \emph{Proceedings of the 2021 ACM conference on fairness,
  accountability, and transparency}, pages 353--362, 2021.

\bibitem[Knothe(1957)]{knothe1957contributions}
H.~Knothe.
\newblock Contributions to the theory of convex bodies.
\newblock \emph{Michigan Mathematical Journal}, 4\penalty0 (1):\penalty0
  39--52, 1957.

\bibitem[Korotin et~al.(2021{\natexlab{a}})Korotin, Egiazarian, Asadulaev,
  Safin, and Burnaev]{korotin2021wasserstein}
A.~Korotin, V.~Egiazarian, A.~Asadulaev, A.~Safin, and E.~Burnaev.
\newblock {W}asserstein-2 generative networks.
\newblock In \emph{International Conference on Learning Representations},
  2021{\natexlab{a}}.

\bibitem[Korotin et~al.(2021{\natexlab{b}})Korotin, Li, Genevay, Solomon,
  Filippov, and Burnaev]{korotin2021neural}
A.~Korotin, L.~Li, A.~Genevay, J.~M. Solomon, A.~Filippov, and E.~Burnaev.
\newblock Do neural optimal transport solvers work? a continuous
  {W}asserstein-2 benchmark.
\newblock \emph{Advances in neural information processing systems},
  34:\penalty0 14593--14605, 2021{\natexlab{b}}.

\bibitem[Korotin et~al.(2023)Korotin, Selikhanovych, and
  Burnaev]{korotin2023neural}
A.~Korotin, D.~Selikhanovych, and E.~Burnaev.
\newblock Neural optimal transport.
\newblock In \emph{The Eleventh International Conference on Learning
  Representations}, 2023.

\bibitem[Kusner et~al.(2017)Kusner, Loftus, Russell, and
  Silva]{kusner2017counterfactual}
M.~J. Kusner, J.~Loftus, C.~Russell, and R.~Silva.
\newblock Counterfactual fairness.
\newblock \emph{Advances in neural information processing systems}, 30, 2017.

\bibitem[Lewis(2013)]{lewis2013counterfactuals}
D.~Lewis.
\newblock \emph{Counterfactuals}.
\newblock John Wiley \& Sons, 2013.

\bibitem[Lin et~al.(2025)Lin, Gao, Blanchet, and Glynn]{lin2025tightening}
S.~Lin, Z.~Gao, J.~Blanchet, and P.~Glynn.
\newblock Tightening causal bounds via covariate-aware optimal transport.
\newblock \emph{arXiv preprint arXiv:2502.01164}, 2025.

\bibitem[Ma and Koenker(2006)]{ma2006quantile}
L.~Ma and R.~Koenker.
\newblock Quantile regression methods for recursive structural equation models.
\newblock \emph{Journal of Econometrics}, 134\penalty0 (2):\penalty0 471--506,
  2006.

\bibitem[Machado et~al.(2025)Machado, Charpentier, and
  Gallic]{machado2025sequential}
A.~F. Machado, A.~Charpentier, and E.~Gallic.
\newblock Sequential conditional transport on probabilistic graphs for
  interpretable counterfactual fairness.
\newblock In \emph{Proceedings of the AAAI Conference on Artificial
  Intelligence}, volume~39, pages 19358--19366, 2025.

\bibitem[MacKillop et~al.(2010)MacKillop, Obasi, Amlung, McGeary, and
  Knopik]{mackillop2010role}
J.~MacKillop, E.~M. Obasi, M.~T. Amlung, J.~E. McGeary, and V.~S. Knopik.
\newblock The role of genetics in nicotine dependence: mapping the pathways
  from genome to syndrome.
\newblock \emph{Current cardiovascular risk reports}, 4:\penalty0 446--453,
  2010.

\bibitem[Makkuva et~al.(2020)Makkuva, Taghvaei, Oh, and
  Lee]{makkuva2020optimal}
A.~Makkuva, A.~Taghvaei, S.~Oh, and J.~Lee.
\newblock Optimal transport mapping via input convex neural networks.
\newblock In \emph{International Conference on Machine Learning}, pages
  6672--6681. PMLR, 2020.

\bibitem[Manole et~al.(2024)Manole, Balakrishnan, Niles-Weed, and
  Wasserman]{manole2024plugin}
T.~Manole, S.~Balakrishnan, J.~Niles-Weed, and L.~Wasserman.
\newblock Plugin estimation of smooth optimal transport maps.
\newblock \emph{The Annals of Statistics}, 52\penalty0 (3):\penalty0 966--998,
  2024.

\bibitem[McCann(1995)]{mccann1995}
R.~J. McCann.
\newblock Existence and uniqueness of monotone measure-preserving maps.
\newblock \emph{Duke Math. J.}, 80\penalty0 (2):\penalty0 309--323, 11 1995.
\newblock \doi{10.1215/S0012-7094-95-08013-2}.

\bibitem[Mesnard et~al.(2021)Mesnard, Weber, Viola, Thakoor, Saade,
  Harutyunyan, Dabney, Stepleton, Heess, Guez,
  et~al.]{mesnard2021counterfactual}
T.~Mesnard, T.~Weber, F.~Viola, S.~Thakoor, A.~Saade, A.~Harutyunyan,
  W.~Dabney, T.~S. Stepleton, N.~Heess, A.~Guez, et~al.
\newblock Counterfactual credit assignment in model-free reinforcement
  learning.
\newblock In \emph{International Conference on Machine Learning}, pages
  7654--7664. PMLR, 2021.

\bibitem[Monge(1781)]{monge1781memoire}
G.~Monge.
\newblock M{\'e}moire sur la th{\'e}orie des d{\'e}blais et des remblais.
\newblock \emph{Histoire de l'Acad{\'e}mie Royale des Sciences de Paris}, 1781.

\bibitem[Mueller and Pearl(2023)]{mueller2023personalized}
S.~Mueller and J.~Pearl.
\newblock Personalized decision making--a conceptual introduction.
\newblock \emph{Journal of Causal Inference}, 11\penalty0 (1):\penalty0
  20220050, 2023.

\bibitem[Nasr-Esfahany et~al.(2023)Nasr-Esfahany, Alizadeh, and
  Shah]{nasr2023counterfactual}
A.~Nasr-Esfahany, M.~Alizadeh, and D.~Shah.
\newblock Counterfactual identifiability of bijective causal models.
\newblock In \emph{International conference on machine learning}, pages
  25733--25754. PMLR, 2023.

\bibitem[Neyman(1923)]{neyman1923applications}
J.~Neyman.
\newblock Sur les applications de la thar des probabilities aux experiences
  agaricales: Essay des principle. excerpts reprinted (1990) in english.
\newblock \emph{Statistical Science}, 5\penalty0 (463-472):\penalty0 4, 1923.

\bibitem[Pascale et~al.(2024)Pascale, Kausamo, and
  Wyczesany]{depascale202460yearscyclicmonotonicity}
L.~D. Pascale, A.~Kausamo, and K.~Wyczesany.
\newblock 60 years of cyclic monotonicity: a survey, 2024.
\newblock URL \url{https://arxiv.org/abs/2308.07682}.

\bibitem[Pearl(2009)]{pearl2009causality}
J.~Pearl.
\newblock \emph{Causality}.
\newblock Cambridge university press, 2009.

\bibitem[Pearl(2010)]{pearl2010brief}
J.~Pearl.
\newblock Brief report: On the consistency rule in causal inference:" axiom,
  definition, assumption, or theorem?".
\newblock \emph{Epidemiology}, pages 872--875, 2010.

\bibitem[Pearl and Mackenzie(2018)]{pearl2018book}
J.~Pearl and D.~Mackenzie.
\newblock \emph{The book of why: the new science of cause and effect}.
\newblock Basic books, 2018.

\bibitem[Peters et~al.(2017)Peters, Janzing, and
  Sch{\"o}lkopf]{peters2017elements}
J.~Peters, D.~Janzing, and B.~Sch{\"o}lkopf.
\newblock \emph{Elements of causal inference: foundations and learning
  algorithms}.
\newblock The MIT press, 2017.

\bibitem[Peterson et~al.(2021)Peterson, Nieto, Wyser, Lambercy, Gassert,
  Milone, and Spies]{peterson2021transfer}
V.~Peterson, N.~Nieto, D.~Wyser, O.~Lambercy, R.~Gassert, D.~H. Milone, and
  R.~D. Spies.
\newblock Transfer learning based on optimal transport for motor imagery
  brain-computer interfaces.
\newblock \emph{IEEE Transactions on Biomedical Engineering}, 69\penalty0
  (2):\penalty0 807--817, 2021.

\bibitem[Peyré and Cuturi(2020)]{peyré2020computationaloptimaltransport}
G.~Peyré and M.~Cuturi.
\newblock Computational optimal transport, 2020.
\newblock URL \url{https://arxiv.org/abs/1803.00567}.

\bibitem[Ple{\v{c}}ko and Meinshausen(2020)]{plevcko2020fair}
D.~Ple{\v{c}}ko and N.~Meinshausen.
\newblock Fair data adaptation with quantile preservation.
\newblock \emph{Journal of Machine Learning Research}, 21\penalty0
  (242):\penalty0 1--44, 2020.

\bibitem[Pooladian and Niles-Weed(2021)]{pooladian2021entropic}
A.-A. Pooladian and J.~Niles-Weed.
\newblock Entropic estimation of optimal transport maps.
\newblock \emph{arXiv preprint arXiv:2109.12004}, 2021.

\bibitem[Redko et~al.(2017)Redko, Habrard, and Sebban]{redko2017theoretical}
I.~Redko, A.~Habrard, and M.~Sebban.
\newblock Theoretical analysis of domain adaptation with optimal transport.
\newblock In \emph{Joint European Conference on Machine Learning and Knowledge
  Discovery in Databases}, pages 737--753. Springer, 2017.

\bibitem[Richens et~al.(2022)Richens, Beard, and
  Thompson]{richens2022counterfactual}
J.~Richens, R.~Beard, and D.~H. Thompson.
\newblock Counterfactual harm.
\newblock \emph{Advances in Neural Information Processing Systems},
  35:\penalty0 36350--36365, 2022.

\bibitem[Rockafellar(1997)]{rockafellar1997convex}
R.~Rockafellar.
\newblock \emph{Convex Analysis}.
\newblock Princeton Landmarks in Mathematics and Physics. Princeton University
  Press, 1997.
\newblock ISBN 9780691015866.
\newblock URL \url{https://books.google.fr/books?id=1TiOka9bx3sC}.

\bibitem[Rosenblatt(1952)]{rosenblatt1952remarks}
M.~Rosenblatt.
\newblock Remarks on a multivariate transformation.
\newblock 23\penalty0 (3):\penalty0 470--472, 1952.
\newblock ISSN 0003-4851.
\newblock \doi{10.1214/aoms/1177729394}.
\newblock URL \url{http://projecteuclid.org/euclid.aoms/1177729394}.
\newblock Publisher: Institute of Mathematical Statistics.

\bibitem[Rubin(1974)]{rubin1974estimating}
D.~B. Rubin.
\newblock Estimating causal effects of treatments in randomized and
  nonrandomized studies.
\newblock \emph{Journal of educational Psychology}, 66\penalty0 (5):\penalty0
  688, 1974.

\bibitem[Sarvet and Stensrud(2025)]{sarvet2025perspectives}
A.~L. Sarvet and M.~J. Stensrud.
\newblock Perspectives on “harm” in personalized medicine.
\newblock \emph{American Journal of Epidemiology}, 194\penalty0 (6):\penalty0
  1743--1748, 2025.

\bibitem[Serfling(2002)]{serfling02}
R.~Serfling.
\newblock Quantile functions for multivariate analysis: approaches and
  applications.
\newblock \emph{Statistica Neerlandica}, 56\penalty0 (2):\penalty0 214--232,
  2002.
\newblock \doi{https://doi.org/10.1111/1467-9574.00195}.
\newblock URL
  \url{https://onlinelibrary.wiley.com/doi/abs/10.1111/1467-9574.00195}.

\bibitem[Souslin(1917)]{souslin1917}
M.~Souslin.
\newblock Sur une definition des ensembles mesurables {B} sans nombres
  transfinis.
\newblock \emph{Comptes rendus de l'Academie des {Sciences} de {Paris}},
  164:\penalty0 88--91, 1917.

\bibitem[Thurin(2024)]{thurin24these}
G.~Thurin.
\newblock \emph{{Multivariate quantiles and regularized optimal transport}}.
\newblock Theses, {Universit{\'e} de Bordeaux}, Nov. 2024.
\newblock URL \url{https://theses.hal.science/tel-04819731}.

\bibitem[Torous et~al.(2024)Torous, Gunsilius, and Rigollet]{torous2024optimal}
W.~Torous, F.~Gunsilius, and P.~Rigollet.
\newblock An optimal transport approach to estimating causal effects via
  nonlinear difference-in-differences.
\newblock \emph{Journal of Causal Inference}, 12\penalty0 (1):\penalty0
  20230004, 2024.

\bibitem[Tu et~al.(2022)Tu, Zhang, Kjellstr{\"o}m, and Zhang]{tu2022optimal}
R.~Tu, K.~Zhang, H.~Kjellstr{\"o}m, and C.~Zhang.
\newblock Optimal transport for causal discovery.
\newblock \emph{International Conference on Learning Representations}, 2022.

\bibitem[Villani(2008)]{villani2008optimal}
C.~Villani.
\newblock \emph{Optimal Transport: Old and New}.
\newblock Grundlehren der mathematischen Wissenschaften. Springer Berlin
  Heidelberg, 2008.
\newblock ISBN 9783540710509.
\newblock URL \url{https://books.google.fr/books?id=hV8o5R7_5tkC}.

\bibitem[Xu et~al.(2020)Xu, Liu, Wang, Chen, and Wang]{xu2020reliable}
R.~Xu, P.~Liu, L.~Wang, C.~Chen, and J.~Wang.
\newblock Reliable weighted optimal transport for unsupervised domain
  adaptation.
\newblock In \emph{Proceedings of the IEEE/CVF conference on computer vision
  and pattern recognition}, pages 4394--4403, 2020.

\end{thebibliography}

\newpage

\appendix

\section{Proofs of Section 2}

\begin{proof}[Proof of \cref{lem:pushforward}] Proof of $(i)$: since $T_1,T_2$ are Borel functions, $T_1 \circ T_2$ is also a Borel function, and for all Borel set $B$,
$$ T_1 \sharp (T_2 \sharp \mu)(B) = T_2 \sharp \mu (T_1^{-1}(B)) = \mu (T_2^{-1}(T_1^{-1}(B))) =  \mu ((T_1 \circ T_2)^{-1}(B))) =  T_1 \circ T_2) \sharp \mu (B) \, .$$
Proof of $(ii)$: if $T_1$ is bijective, then since $\R^d$ is a Polish space $T_1^{-1}$ is also Borel, via Souslin's theorem (\cite{souslin1917}), and
\begin{align*}
    \nu = T_1 \sharp \mu & \iff  \mbox{for all Borel set } B, \; \mu (T_1^{-1}(B)) = \nu(B) \\
    & \iff \mbox{for all Borel set } C, \; \mu (C) = \mu(T_1^{-1}(T_1(C)) = \nu(T_1(C)) = \nu((T_1^{-1})^{-1}(C)) \\
    & \iff \mu = T_1^{-1} \sharp \nu.
\end{align*}
\end{proof}

\begin{proof}[Proof of \cref{lem:CM_is_M_in_1D}]
First, if $T: \R \to \R$ is a cyclically monotone map, then for all $x,y \in \R$, $0 \leq x(T(x)-T(y)) + y(T(y)-T(x)) = (x-y)(T(x)-T(y))$, that is $T$ is non-decreasing. Conversely, if $T : \R \to \R$ is non-decreasing, then for all $x^{(1)}, \ldots, x^{(m)}$ the rearrangement inequality yields
$ \sum_{i=1}^{m} x^{(i)} T(x^{(i)}) \geq \sum_{i=1}^{m} x^{(i+1)} T(x^{(i)}), $ hence the cyclic monotonicity of $T$.
\end{proof}

\begin{proof}[Proof of \cref{prop:OT_invertibility}]
By \cref{thm:mccann}, there exists a closed, proper convex function $\varphi: \R^d \to \R \cup \{ + \infty \}$ such that $\CM(\mu,\nu) = \nabla \varphi$, $\mu$-almost everywhere. The convex (Fenchel-Legendre) conjugate of $\varphi$ defined by
$$ \forall y \in \R^d \, , \quad \varphi^*(y) := \sup_{x \in \R^d} (\langle y,x \rangle - \varphi(x))$$ is also a closed, proper convex function from $\R^d$ to $\R \cup \{ + \infty \}$. Moreover, since $\varphi$ is closed and proper, Corollary 23.5.1. in \cite{rockafellar1997convex} implies that $x \in \partial \varphi^*(y) \iff y \in \partial \varphi(x)$ for all $x,y \in \R^d$. The above, together with the fact that $\nu$ also has a density, implies that $\varphi^*$ is $\nu-$a.e. differentiable, and $\nabla \varphi^* \circ \nabla \varphi (x) = x$ $\mu-$a.e, and $\nabla \varphi \circ \nabla \varphi^* (y) = y$ $\nu-$a.e. 

Now, for all Borel set $B$, $\nu([\nabla\varphi^*]^{-1}(B)) = \nu([\nabla\varphi](B)) = \mu([\nabla\varphi]^{-1}([\nabla\varphi](B))) = \mu(B) $, where we used the fact that $\nabla\varphi \in \T(\mu,\nu)$. This shows that $\nabla\varphi^* \in \T(\nu,\mu)$, and since $\varphi^*$ is also a closed, proper convex function, another appeal to \cref{thm:mccann} yields $\nabla\varphi^* = \CM(\nu,\mu)$ almost everywhere, which concludes the proof.
\end{proof}

\begin{proof}[Proof of \cref{thm:QPmaps}]
    The fact that $\QP(\mu,\nu \, ; \, p_0)$ satisfies \eqref{eq:p0-QP} and its uniqueness $\mu-$a.e. directly follow from \cref{prop:OT_invertibility}.
\end{proof}

\begin{proof}[Proof of \cref{lem:QP_in_1D}]
    Let $\square$ denote the uniform measure on $[0,1]$. Denote by $F_0$ the c.d.f. of $p_0$. It is clear that $F_\nu^{\dagger} \circ F_0 = \CM(\square,\nu) \circ \CM(p_0,\square)$ is a transport map from $p_0$ to $\nu$. Moreover, it is CM (since it is non decreasing and $d=1$), and $p_0$ has a density. Thus, by \cref{thm:mccann}, $F_\nu^{\dagger} \circ F_0 = \CM(p_0,\nu)$ a.e. With the same arguments, $\CM(\mu,p_0) = F_0^{\dagger} \circ F_\mu$ a.e. which gives a.e.,
    $$ \CM(p_0,\nu) \circ \CM(\mu,p_0) = F_\nu^{\dagger} \circ F_0 \circ F_0^{\dagger} \circ F_\mu = F_\nu^{\dagger} \circ F_\mu, $$ where we used that $F_0 \circ F_0^{\dagger}= \Id$ since $p_0$ is atomless ($F_0$ is continuous).  
\end{proof}

\begin{proof}[Proof of \cref{lem:1Dcase}]
    It is trivial to check that $F^{\dagger}_{\nu} \circ F_{\mu} \in \T(\mu,\nu)$. Writing every equality $\mu-$almost everywhere, since $F_{\mu}$ and $F^{\dagger}_{\nu}$ are non-decreasing, then so is $F^{\dagger}_{\nu} \circ F_{\mu}$ and thus $\CM(\mu,\nu) = F^{\dagger}_{\nu} \circ F_{\mu}$. \cref{lem:QP_in_1D} shows that $\QP(\mu,\nu ; p_0) = F^{\dagger}_{\nu} \circ F_{\mu}$. This same map is in turn triangular monotone ($d=1$) and uniqueness of the KR map yields $\KR(\mu,\nu) = F^{\dagger}_{\nu} \circ F_{\mu}$.
\end{proof}

\begin{proof}[Proof of \cref{lem:productcase}]
    Since $T$ is diagonal and the $F^{\dagger}_{\nu_i} \circ F_{\mu_i}$ are non-decreasing, $T$ is TM. 
    Moreover, $T$ is $\mu-$a.e. the gradient of the convex function $x \mapsto \int_{[0,x_1]} F^{\dagger}_{\nu_1} \circ F_{\mu_1}(t_1)\rmd t_1 + \ldots + \int_{[0,x_d]} F^{\dagger}_{\nu_d} \circ F_{\mu_d}(t_d)\rmd t_d$, hence it is CM. By \cref{thm:mccann} and uniqueness of the KR map, $\mu-$a.e.,
$\KR(\mu,\nu) = \CM(\mu,\nu) = T.$ 
    Now, assume that $p_0$ writes $p_0 = \otimes_{i=1}^{d} p_{0,i}$ where the $p_{0,i}$ have densities, with c.d.f.s $F_{p_{0,i}}$. Thanks to the first part of the proof, we know that $\mu-$a.e.,
    $$\CM(\mu,p_0) = \left(F^{\dagger}_{p_{0,1}} \circ F_{\mu_1}, \ldots, F^{\dagger}_{p_{0,d}} \circ F_{\mu_d} \right)$$ and $p_0-$a.e.,
    $$\CM(p_0,\nu) = \left(F^{\dagger}_{\nu_1} \circ F_{p_{0,1}}, \ldots, F^{\dagger}_{\nu_d} \circ F_{p_{0,d}} \right) \, .$$ By, definition, $\mu-$a.e. $\QP(\mu,\nu;p_0) = \CM(p_0,\nu) \circ \CM(p_0,\mu)^{-1}$. By \cref{prop:OT_invertibility}, $\mu-$a.e.,
    \begin{align*}
        \QP(\mu,\nu;p_0) & = \CM(p_0,\nu) \circ \CM(p_0,\mu)^{-1} \\ 
        & = \CM(p_0,\nu) \circ \CM(\mu,p_0) \\
        & = \left(F^{\dagger}_{\nu_1} \circ F_{p_{0,1}} \circ F^{\dagger}_{p_{0,1}} \circ F_{\mu_1}, \ldots, F^{\dagger}_{\nu_d} \circ F_{p_{0,d}} \circ F^{\dagger}_{p_{0,d}} \circ F_{\mu_d}\right) \\
        & = \left(F^{\dagger}_{\nu_1} \circ F_{\mu_1}, \ldots, F^{\dagger}_{\nu_d} \circ F_{\mu_d}\right) \\
        & = T, 
    \end{align*}
    where we used that all $F_{p_{0,i}} \circ F_{p_{0,i}}^{\dagger}= \Id$ since $p_{0,i}$ is atomless ($F_{p_{0,i}}$ is continuous). This concludes the proof.
\end{proof}

\begin{proof}[Proof of \cref{prop:TM-CM_maps}]
First, let us focus on $C$, identified with $\CM(\mu,C\sharp\mu)$. By \cref{thm:mccann}, and by definition of the CM map, $C$ is the gradient of a proper convex function $\varphi : \R^d \to \R \cup \{ + \infty\}$. The convexity of $\varphi$ yields, by Alexandrov’s second differentiability theorem (see e.g. \citep[Theorem 14.24]{villani2008optimal}) that $\varphi$ is Lebesgue-a.e., thus $\mu-$a.e., twice differentiable. Differentiating $C = \nabla \varphi$ gives that $\mu-$a.e,
$$ \mathrm{Jac}(C) = \nabla^2 \varphi, $$ which implies that $\mu-$a.e, $\mathrm{Jac}(C)$ is symmetric positive semi-definite (since it is the Hessian of a convex function). 

Now recall that we assume that $\mu-$a.e., $C = T$.
The intuition is as follows: since $C$ agrees $\mu-$a.e. with a triangular map, which Jacobian -- if it exists -- must be a lower triangular matrix, then $\mathrm{Jac}(C)$ is also lower triangular $\mu-$a.e. 

This previous rationale is incorrect, because $C = T$ $\mu-$a.e. does not imply that $T$ is differentiable $\mu-$a.e. (differentiability is not preserved by $\mu-$null sets perturbations). 
Let us nonetheless prove by contradiction that our intuition is correct. Assume that there exists $1 \leq i<j \leq d$ and a set $\dE \subseteq \R^d$ with $\mu(\dE)>0$ on which 
$$ \mbox{for all } x \in \dE, \quad \mathrm{Jac}(C)(x)_{i,j} \neq 0 \, .$$ Take such an $x \in \dE$. Since $\mathrm{Jac}(C)(x)_{i,j} = \frac{\partial C_i}{\partial x_j}(x) \neq 0$, there is a real number $u \neq x_i$, close enough to $x_i$ such that $C_i(x_1, \ldots, x_{j-1},u,x_{j+1}, \ldots, x_{d})$ and $C_i(x)$ are different. Denoting
$$\mathcal{F} := \{ x \in \R^d \, | \, \exists u \neq x_j, C_i(x_1, \ldots, x_{j-1},u,x_{j+1}, \ldots, x_{d}) \neq C_i(x)\},
$$
we have that $\mathcal{F}$ is a Borel set and we just proved $\mathcal{F} \supsetneq \dE$. Thus $\mu(\mathcal{F}) \geq \mu(\dE) >0$, which contradicts the fact that $C$ is $\mu-$a.e. triangular. The $\mu-$a.e. lower triangular structure of $\mathrm{Jac}(C)$ together with the fact that $\mathrm{Jac}(C)$ is $\mu-$a.e. symmetric positive semi-definite gives that $\mu-$a.e., $\mathrm{Jac}(C)$ is diagonal with positive coefficients. 

Thus, because $C$ is differentiable $\mu-$a.e., then by Lebesgue integration, $C$ is diagonal non-decreasing $\mu-$a.e., which concludes the proof.
\end{proof}

\begin{proof}[Proof of \cref{cor:CM=KR}]
The proof of $(ii) \implies (i)$ directly follows from the observation that diagonal non-decreasing maps are both CM and TM Lebesgue-a.e., thus $\mu-$a.e. $(i) \implies (ii)$ is a direct consequence of \cref{prop:TM-CM_maps}.
\end{proof}

\begin{proof}[Proof of \cref{lem:reference}]
Identity is straightforward: $T_{a \leftarrow a} = g_a \circ g^{-1}_a = \Id$. For path independence, specifying the right-hand side of the equality with the expression of the transport maps gives
\[
    g_{a''} \circ g^{-1}_{a'} \circ g_{a'} \circ g^{-1}_a = g_{a''} \circ g^{-1}_a = T_{a'' \leftarrow a}. 
\]
To prove inversion, we apply path independence with $a''=a$, so that, according to identity, $\Id = T_{a \leftarrow a'} \circ T_{a' \leftarrow a} = T_{a' \leftarrow a} \circ T_{a \leftarrow a'} $.
\end{proof}

\begin{proof}[Proof of \cref{prop:algebraic_classic}]
We start with $(i)$. Identity simply follows from $P_a = \Id \sharp P_a$ and $\Id$ being the gradient of a convex function, which implies $\Id = \CM(P_a,P_a)$. Inversion is a direct consequence of \cref{prop:OT_invertibility}. To show that path independence does not hold in general, consider the following counterexample, where $\A = \{ 0,1,2 \}$. Let $T_{1 \leftarrow 0} : (x,y) \mapsto (3x, y)$, which is clearly CM, $\varphi_{2 \leftarrow 1} : (x,y) \mapsto e^{x + y^2} + x^2$, and  
$T_{2 \leftarrow 1}(x,y) = \nabla \varphi_{2 \leftarrow 1}(x,y) = \left( 2x + e^{x + y^2},\ 2y e^{x + y^2} \right)$, which is also CM since $\varphi_{2 \leftarrow 1}$ is convex.  Consider $P_0 = \Unif([0,1])^{\otimes 2}$, $P_1 = T_{1 \leftarrow 0} \sharp P_0$, and $P_2 = T_{2 \leftarrow 1} \sharp P_1$. By definition, $T_{1 \leftarrow 0} = \CM(P_0, P_1)$ and $T_{2 \leftarrow 1} = \CM(P_1, P_2)$. Note however that $T_{2 \leftarrow 1} \circ T_{1 \leftarrow 0}$, given by  
$$
T_{2 \leftarrow 1} \circ T_{1 \leftarrow 0} (x,y) = \left( 6x + e^{3x + y^2},\ 2y e^{3x + y^2} \right),
$$  
is not the gradient of any function (convex or not).  
This can be clearly seen by differentiating the first component with respect to $y$, and the second component with respect to $x$: we obtain different results. Thus, $T_{2 \leftarrow 1} \circ T_{1 \leftarrow 0} \neq \CM(P_0,P_2)$.

Next, $(ii)$ follows from the definition $\QP(P_a,P_{a'};p_0) = \CM(p_0,P_{a'}) \circ \CM(p_0,P_a)^{-1}$. The $\QP(P_a,P_{a'};p_0)$ are reference-based transport maps, and we conclude with \cref{lem:reference}.

We turn to the proof of $(iii)$. Identity is a direct consequence of $\Id$ being a TM map from $P_a$ to $P_a$. According to \cref{lem:pushforward}, $\KR(P_{a'},P_{a''}) \circ \KR(P_a,P_{a'})$ is a transport map from $P_a$ to $P_{a''}$. Furthermore, being TM is stable under inversion and since $\KR(P_{a'},P_{a})^{-1}$ is a transport map from $P_{a'}$ to $P_a$ according to Lemma~\ref{lem:pushforward}, $\KR(P_{a'},P_{a})^{-1} = \KR(P_a,P_{a'})$ by uniqueness. Similarly, being TM is stable under composition, therefore, $\KR(P_{a'},P_{a''}) \circ \KR(P_a,P_{a'})$ is a TM transport map from $P_a$ to $P_{a''}$. It then follows from uniqueness that $\KR(P_{a'},P_{a''}) \circ \KR(P_a,P_{a'}) = \KR(P_a,P_{a''})$. This shows path independence. 
\end{proof}

\section{Proofs of Section 3}

\begin{proof}[Proof of \cref{prop:submap}]
Let $S_{\ttx} : \A_{\pa(\ttx)} \times \U_{\ttx} \to \X$ be the solution map of $\M$ with respect to $[d]$ and let $a \in \A$. Then, note that $S^{(a)} : \U \to \X \times \A, u \mapsto (S_{\ttx}(a_{\pa([d])},u_{\ttx}),a)$ is the solution map of $\M_a$. This means that $\M_a$ is uniquely solvable. Additionally, for every solution $(X_a,a,U)$ of $\M_a$, we have $(X_a,a) = S^{(a)}(U)$ by definition of a solution map. This leads to $X_a = S_{\ttx}(a_{\pa([d])},U_{\ttx})$. Defining $g_a : \U_{\ttx} \to \X, u_{\ttx} \mapsto S_{\ttx}(a_{\pa([d])},u_{\ttx})$ concludes the proof.
\end{proof}

\begin{proof}[Proof of \cref{prop:algebraic_ctf}]
It is a direct consequence of \cref{lem:reference} and the expression of counterfactual maps.
\end{proof}

\begin{proof}[Proof of \cref{prop:reparameterization}]
First, we show that $\widetilde{\M} \in \SCM$ is uniquely solvable. Since $\M \in \SCM$, there exists a measurable map $S$ such that, for $Q$-almost every $u$ and every $v \in \V$,
\[
    v = S(u) \iff v_i = f_i(v_{\pa(i)},u_i) \text{ for every } i \in \I.    
\]
Using the fact that $Q = (\times_{i \in \I} \phi_i) \sharp \widetilde{Q}$, this equivalence can be reframed as: for $\widetilde{Q}$-almost every $\widetilde{u}$ and every $v \in \V$,
\[
    v = S((\times_{i \in \I} \phi_i)(\widetilde{u})) \iff v_i = f_i(v_{\pa(i)},\phi_i(\widetilde{u_i})) \text{ for every } i \in \I.    
\]
Noting that $f_i(v_{\pa(i)},\phi_i(\widetilde{u_i})) = \widetilde{f}_i(v_{\pa(i)},\widetilde{u_i})$, this proves by definition that $\widetilde{\M}$ is uniquely solvable with respect to $\I$ such that $\widetilde{S} = S \circ (\times_{i \in \I} \phi_i)$.

Similarly, we show that $\widetilde{\M} \in \SCM$ is uniquely solvable with respect to $[d]$. Since $\M \in \SCM$, there exists a measurable map $S_{\ttx}$ such that, for $Q_{\ttx}$-almost every $u_{\ttx}$ and every $(x,a) \in \X \times \A$,
\[
    x = S_{\ttx}(a_{\pa([d])},u_{\ttx}) \iff x_i = f_i(a_{\pa(i)}, x_{\pa(i)},u_i) \text{ for every } i \in [d].
\]
Using the fact that $Q_{\ttx} = (\times^d_{i=1} \phi_i) \sharp \widetilde{Q}_{\ttx}$, this equivalence can be reframed as: for $\widetilde{Q}_{\ttx}$-almost every $\widetilde{u}_{\ttx}$ and every $(x,a) \in \X \in \A$,
\[
    x = S_{\ttx}(a_{\pa([d])},(\times^d_{i=1} \phi_i)(\widetilde{u}_{\ttx})) \iff x_i = f_i(a_{\pa(i)}, x_{\pa(i)},\phi_i(\widetilde{u}_i)) \text{ for every } i \in [d].
\]
Noting that $\widetilde{f}_i(a_{\pa(i)}, x_{\pa(i)},\widetilde{u}_i) = f_i(a_{\pa(i)}, x_{\pa(i)},\phi_i(\widetilde{u}_i))$, this proves by definition that $\widetilde{\M}$ is uniquely solvable with respect to $[d]$ such that $\widetilde{S}_{\ttx}(a_{\pa([d])},\widetilde{u}_{\ttx}) = S_{\ttx}(a_{\pa([d])},(\times^d_{i=1} \phi_i)(\widetilde{u}_{\ttx}))$ for every $(a,\widetilde{u}) \in \A \times \widetilde{\U}$.

Second, we provide expressions for the subsolution maps and the counterfactual maps of $\widetilde{\M}$. Let $a \in \A$. According to \cref{prop:submap}, $g_a(u_{\ttx}) := S_{\ttx}(a_{\pa([d])},u_{\ttx})$ and $\widetilde{g}_a(\widetilde{u}_{\ttx}) := \widetilde{S}_{\ttx}(a_{\pa([d])},\widetilde{u}_{\ttx})$. Using the fact that $\widetilde{S}_{\ttx}(a_{\pa([d])},\widetilde{u}_{\ttx}) = S_{\ttx}(a_{\pa([d])},(\times^d_{i=1} \phi_i)(\widetilde{u}_{\ttx}))$, we obtain $\widetilde{g}_a = g_a \circ (\times^d_{i=1} \phi_i)$. Next, let $a,a' \in \A$ and compute,
\begin{align*}
    \widetilde{C}_{a' \leftarrow a} &= \widetilde{g}_{a'} \circ \widetilde{g}^{-1}_a\\ &= g_{a'} \circ (\times^d_{i=1} \phi_i) \circ \left(g_a \circ (\times^d_{i=1} \phi_i)\right)^{-1}\\ &= g_{a'} \circ (\times^d_{i=1} \phi_i) \circ (\times^d_{i=1} \phi_i)^{-1} \circ g^{-1}_a\\ &= g_{a'} \circ g^{-1}_a\\ &= C_{a' \leftarrow a}.
\end{align*}
This concludes the proof.
\end{proof}

\begin{proof}[Proof of \cref{prop:acyclic}]
The fact that an acyclic SCM is uniquely solvable with respect to any subset of variables follows from \citep[Proposition 3.4]{bongers2021foundations}.

Let $a \in \A$ and $i \in [d]$. One determines $g_a$ by solving the system of equations $\{ x_i = f_i(a_{\pa(i)},x_{\pa(i)},u_i) : i \in [d] \}$ with unknown variable $x \in \X$. Since $\M$ is acyclic, it induces a topological order on $[d]$. Recursive substitutions of the $x_i$ by $f_i(a_{\pa(i)},x_{\pa(i)},u_i)$ along this order yield the expression
\[
    g_a(u_{\ttx})_i = f_i(a_{\pa(i)},\psi_i(a,u_{\an(i) \setminus \{i\}}),u_i)
\]
for some measurable map $\psi_i$.

Determining $g^{-1}_a$ amounts to solving the system of equations $\{ x_i = g_a(u_{\ttx})_i : i \in [d] \}$ with unknown variable $u_{\ttx} \in \U_{\ttx}$. Using the expression of $g_a$ and the injectivity of $f_i$ in its exogenous variable, one has $u_i = f_i(a_{\pa(i)},\phi_i(a,u_{\an(i) \setminus \{i\}}),\cdot)^{-1}(x_i)$. Again, using recursive substitution along the topological order gives
\[
    g^{-1}_a(x)_i = f_i(a_{\pa(i)},\tilde{\psi}_i(a,x_{\an(i) \setminus \{i\}}),\cdot)^{-1}(x_i)
\]
for some measurable map $\tilde{\psi}_i$.

Let $a,a' \in \A$ and $i \in [d]$. By definition, and using the previous expression,
\[
    C_{a' \leftarrow a}(x)_i = f_i(a'_{\pa(i)},\psi_i(a',g^{-1}_a(x)_{\an(i) \setminus \{i\}}),g^{-1}_a(x)_i).
\]
To conclude, note that $g_a$, $g^{-1}_a$, and $C_{a' \leftarrow a}$ only depend on their inputs indexed by $\an(i)$. Therefore, they are lower triangular with respect to the \emph{same} permutation.
\end{proof}

\begin{proof}[Proof of \cref{lem:linear}]
It holds that, $v_i = f_i(v_{\pa(i)},u_i)$ for every $i \in [d]$ $\iff$ $v_I = M_{I,I} v_I + M_{I,\pa(I) \setminus I} v_{\pa(I) \setminus I} + b_I + u_I$, by concatenation and using the fact that $M_{i,j} = 0 \iff j \notin \pa(i)$. This equation is solvable in $v_I$ if and only if $\Id - M_{I,I}$ is invertible. If it is the case, we obtain the solution $v_I = (\Id - M_{I,I})^{-1}(M_{I,\pa(I) \setminus I} v_{\pa(I) \setminus I} + b_I + u_I)$. By definition, the solution map has the proposed expression.
\end{proof}

\begin{proof}[Proof of \cref{prop:ctf_linear}]
The model is uniquely solvable with respect to $[d]$ by assumption. Then, the expression of $g_a$ directly follows from \cref{lem:linear}. To compute the counterfactual map, note that $g^{-1}_a(x) = (\Id - M^{\ttx}) x - m^{\ttx} a - b^{\ttx}$. This leads to
\begin{align*}
    C_{a' \leftarrow a}(x) &= (\Id - M^{\ttx})^{-1}\left(m^{\ttx} a' + b^{\ttx} + (\Id - M^{\ttx}) x - m^{\ttx} a - b^{\ttx}\right)\\
    &= x + (\Id - M^{\ttx})^{-1} m^{\ttx} (a' - a).
\end{align*}
\end{proof}

\section{Proofs of Section 4}

\begin{proof}[Proof of \cref{lem:markov}]
\emph{Item $(i)$.} Suppose that $(P_a)_{a \in \A}$ is $\SCM^{\G}_{acy}-$compatible: there exists $\M \in \SCM_{acy}$ with graph $\G$ and with $(P_a)_{a \in \A}$ as interventional marginals.

First of all, let $a \in \A$ and write $(X_a,a,U)$ for a solution of $\M_a$. Recall that $X_a \sim P_a$. As such, using the chain rule,
\[
    \rmd P_a(x_1,\ldots,x_d) = \rmd \bbP(X_a=(x_1,\ldots,x_d)) = \prod^d_{i=1} \rmd \bbP(X_{a,i}=x_i | X_{a,[i-1]} = x_{[i-1]}).
\]
Next, using the SCM representation, note that $X_{a,i} = f_i(a_{\pa(i)},X_{a,\pa(i)},U_i)$ for every $i \in [d]$. We rely on two properties. First, $X_{a,[i-1]} \independent U_i$, due to acyclicity and the independence of the exogenous noises (see for instance \citep[Remark 3.2]{cheridito2025optimal}). Second, $\{1,\ldots,d\}$ is topologically ordered. Therefore,
\begin{align*}
    \rmd \bbP(X_a=x) &= \prod^d_{i=1} \rmd \bbP(X_{a,i} = x_i | X_{a,[i-1]} = x_{[i-1]}) \text{ by the chain rule}\\
    &= \prod^d_{i=1} \rmd \bbP(f_i(a_{\pa(i)},X_{a,\pa(i)},U_i) = x_i | X_{a,[i-1]} = x_{[i-1]}) \text{ using the SCM}\\ &= \prod^d_{i=1} \rmd \bbP(f_i(a_{\pa(i)},x_{a,\pa(i)},U_i) = x_i | X_{a,[i-1]} = x_{[i-1]}) \text{ since } \pa(i) \subseteq [i-1]\\
    &= \prod^d_{i=1} \rmd \bbP(f_i(a_{\pa(i)},x_{a,\pa(i)},U_i) = x_i | X_{a,\pa(i)} = x_{\pa(i)}) \text{ since } X_{a,[i-1]} \independent U_i\\
    &= \prod^d_{i=1} \rmd \bbP(f_i(a_{\pa(i)},X_{a,\pa(i)},U_i) = x_i | X_{a,\pa(i)} = x_{\pa(i)})\\
    &= \prod^d_{i=1} \rmd \bbP(X_{a,i} = x_i | X_{a,\pa(i)} = x_{\pa(i)}).
\end{align*}

\emph{Item $(ii)$.} We introduce an extra notation. For every $a \in \A$, every $i \in [d]$, and every $x_{[i-1]} \in \X_{[i-1]}$, we denote by $P_{a,i}(\cdot|x_{[i-1]})$ the distribution under $P_a$ of the $i$th variable conditional to the $i-1$ previous variables being valued at $x_{[i-1]}$. The proof crucially rests on the following factorization:
\[
    \rmd P_a(x_1,\ldots,x_d) = \prod^d_{i-1} \rmd P_{a,i}(\cdot|x_{[i-1]}),
\]
obtained by the chain rule.

Let $\G$ be the graph defined as $\pa(\tta) = \varnothing$ and $\pa(i) = \{\tta\} \cup [i-1]$ for every $i \in [d]$. The demonstration consists in constructing an SCM with graph $\G$ that fits the interventional marginals. In a first time, we focus on designing the causal mechanism. To do so, we rely on \emph{noise outsourcing} \citep[Proposition 8.20]{kallenberg1997foundations}: for any two random variables $R_1$ and $R_2$, there exists a measurable map $h$ and a random variable $E \sim \Unif([0,1])$ such that $R_1 = h(R_2,E)$, where $E \independent R_2$. In particular, $\law{R_1 | R_2 = r_2} = \law{h(r_2,E)} =  h(r_2,\cdot) \# \Unif([0,1])$. This result implies in our case that, for every $a \in \A$ and every $i \in [d]$, there exists a measurable function $h_{a,i} : \X_{[i-1]} \times [0,1] \to \X_i$ such that $P_{a,i|[i-1]}(\cdot|x_{[i-1]}) = h_{a,i}(x_{[i-1]},\cdot) \# \Unif([0,1])$. Then, for each $i \in [d]$, define the function
\[
    f_i : \A \times \X_{[i-1]} \times [0,1] \to \X_i, \quad (a,x_{[i-1]},u_i) \mapsto h_{a,i}(x_{[i-1]},u_i),
\]
which is measurable by countability of $\A$. Finally, let $f_{\tta}$ be any measurable function from $[0,1]$ to $\A$.

Next, consider $\M := (\X \times \A,\G,\U,Q,f)$ over $\I := \{ 1,\ldots,d,\tta \}$, where $\U := [0,1]^{d+1}$, $Q := \Unif([0,1])^{\otimes (d+1)}$, while $\G$ and $f$ are defined as above. In a second time, we show that $\M$ admits $(P_a)_{a \in \A}$ as interventional marginals. Let $a \in \A$ and $(X_a,a,U)$ be a solution of $\M_a$. Then,
\begin{align*}
    \rmd \bbP(X_a=(x_1,\ldots,x_d)) &= \prod^d_{i=1} \rmd \bbP(X_{a,i}=x_i | X_{a,[i-1]} = x_{[i-1]}) \text{ by the chain rule}\\
    &= \prod^d_{i=1} \rmd \bbP(f_i(a,X_{a,[i-1]},U_i)=x_i | X_{a,[i-1]} = x_{[i-1]}) \text{ by definition of $X_a$}\\
    &= \prod^d_{i=1} \rmd \bbP(h_{a,i}(X_{a,[i-1]},U_i)=x_i | X_{a,[i-1]} = x_{[i-1]}) \text{ by definition of $f_i$}\\
    &= \prod^d_{i=1} \rmd \bbP(h_{a,i}(x_{a,[i-1]},U_i)=x_i) \text{ since } \ X_{a,[i-1]} \independent U_i\\
    &= \prod^d_{i=1} \rmd \left( h_{a,i}(x_{a,[i-1]}, \cdot) \sharp \Unif([0,1]) \right)(x_i)\\
    &= \prod^d_{i=1} \rmd P_{a,i}(x_i|x_{[i-1]}) \text{  by definition of $h_{a,i}$}\\
    &= \rmd P_a(x_1,\ldots,x_d).
\end{align*}
This concludes the proof.
\end{proof}

\begin{proof}[Proof of \cref{thm:main}]
\emph{Item $(i)$.} This is a direct extension of the proof of \citep{torous2024optimal} [Theorem 2] that we reproduce here for completeness. Let $a,a' \in \A$, $m \geq 1$. Then, simply note that,
\[
    \sum^m_{i=1} \langle x^{(i)}, g_{a'}(g^{-1}_a(x^{(i)}) - g_{a'}(g^{-1}_a(x^{(i+1)}))  \rangle \geq 0 \iff \sum^m_{i=1} \langle g_a(u_{\ttx}^{(i)}), g_{a'}(u_{\ttx}^{(i)}) - g_{a'}(u_{\ttx}^{(i+1)})  \rangle \geq 0,
\]
where $u^{(i)}_{\ttx} = g^{-1}_a(x^{(i)})$ for every $i \in [m]$.

\emph{Item $(ii)$.} Let us start with the \say{if} sense. Suppose that there exists an injective map $\Phi \in \T(p_0,Q_{\ttx})$ such that, for every $a \in \A$, $g_a \circ \Phi$ is cyclically monotone. According to \cref{lem:pushforward}, $g_a \circ \Phi$ is a cyclically monotone transport map from $p_0$ to $P_a$. Therefore, by uniqueness, $g_a \circ \Phi = \CM(p_0,P_a)$.  Then, for every $a',a \in \A$, $g_{a'} \circ g^{-1}_a = (g_{a'} \circ \Phi) \circ (\Phi^{-1} \circ g^{-1}_a) = \CM(p_0,P_{a'}) \circ \CM(P_a,p_0) = \QP(P_a,P_{a'};p_0)$. We turn to the \say{only if} sense. Suppose that for every $a,a' \in \A$, $g_{a'} \circ g^{-1}_a = \QP(P_a,P_{a'};p_0) = \CM(p_0,P_{a'}) \circ \CM^{-1}(p_0,P_a)$. Let $a_0 \in \A$ be fixed. Then, $g_a \circ g^{-1}_{a_0} \circ \CM(p_0,P_{a_0}) = \CM(p_0,P_a)$ for every $a \in \A$. To conclude, letting $\Phi := g^{-1}_{a_0} \circ \CM(p_0,P_{a_0})$, $\Phi$ is measurable and invertible by composition, and belongs to $\T(p_0,Q_{\ttx})$. In some of the future proofs, we shall use this characterization in the form: $g_a = \CM(p_0,P_a) \circ \Phi^{-1}$ for every $a \in \A$.
\end{proof}

\begin{proof}[Proof of \cref{thm:main_acy}]
\emph{Item $(i)$.} Let $a,a' \in \A$, and note that $C_{a' \leftarrow a}$ is lower triangular according to \cref{prop:acyclic}. If $C_{a' \leftarrow a} = \CM(P_a,P_{a'})$, then it follows from \cref{prop:TM-CM_maps} that $C_{a' \leftarrow a}$ is diagonal non-decreasing, and from \cref{cor:CM=KR} that $C_{a' \leftarrow a} = \KR(P_a,P_{a'})$. If, conversely, $C_{a' \leftarrow a}$ is diagonal non-decreasing, then $C_{a' \leftarrow a}$ is cyclically monotone.

\emph{Item $(ii)$.} We start with the \say{if} sense. By assumption, for every $a,a' \in \A$, every $i \in [d]$, and every $u_1 \in \U_1,\ldots,u_i \in \U_i,u'_i \in \U_i,\ldots, u_d \in \U_d$,
\[
        (g_a(\ldots,u_i,\ldots)_i-g_a(\ldots,u'_i,\ldots)_i)\\(g_{a'}(\ldots,u_i,\ldots)_i-g_{a'}(\ldots,u'_i,\ldots)_i) \geq 0.
\]
The strategy consists in making a change of variables, based on the invertibility of the $f_i$s and the $g_a$s. Fix $i \in [d]$, and define $x_1 \in \X_i,\ldots,x_i \in \X_i,x'_i \in \X_i,\ldots,x_d \in \X_d$ as follows:
\begin{align*}
    &x_j = f_j(a_{\pa(j)},x_{\pa(j)},u_j) \text{ for every }j \in \an(i) \setminus \{i\},\\
    &x_i = f_i(a_{\pa(i)},x_{\pa(i)},u_i),\\
    &x'_i = f_i(a_{\pa(i)},x_{\pa(i)},u'_i),\\
    &x_j \in \V_j \text{ for every }j \notin \an(i).
\end{align*}
Recall that $[d]$ is topologically ordered. Observe that $(x_1,\ldots,x_{i-1},x_i)$ and $(x_1,\ldots,x_{i-1},x'_i)$ are solutions of $\M$ relative to $[i]$ for, respectively, $(u_1,\ldots,u_{i-1},u_i)$ and $(u_1,\ldots,u_{i-1},u'_i)$ as exogenous inputs. Their first $i-1$ components are equal since the distinct noises $u_i$ and $u'_i$ only serve to define the $i$th variable. For $j \geq i$, we set $x_j$ to any value since $g^{-1}_a$ is constant in these arguments by triangularity. 

As such, by definition of the subsolution map, $x_i = g_a(\ldots,u_i,\ldots)_i$ and $x'_i = g_a(\ldots,u'_i,\ldots)_i$, while $(\ldots,u_i,\ldots) = g^{-1}_a(\ldots,x_i,\ldots)$ and $(\ldots,u'_i,\ldots) = g^{-1}_a(\ldots,x'_i,\ldots)$. Therefore,
\[
        (x_i-x'_i)\\(g_{a'}(g^{-1}_a(\ldots,x_i,\ldots))_i-g_{a'}(g^{-1}_a(\ldots,x'_i,\ldots))_i) \geq 0,
\]
which entails
\[
        (x_i-x'_i)\\(C_{a' \leftarrow a}(\ldots,x_i,\ldots)_i-C_{a' \leftarrow a}(\ldots,x'_i,\ldots)_i) \geq 0.
\]
Thus, $C_{a' \leftarrow a,i}$ is monotone increasing in its $i$th input. Since $C_{a' \leftarrow a}$ is triangular, we conclude that $C_{a' \leftarrow a}$ is TM, and hence $C_{a' \leftarrow a} = \KR(P_a,P_{a'})$. The \say{only if} sense follows from using the inverse change of variables.
\end{proof}

\begin{proof}[Proof of \cref{prop:ctf_linear_CM_KR}]
According to \cref{prop:ctf_linear} $C_{a' \leftarrow a}$, is diagonal non-decreasing. Therefore, $C_{a' \leftarrow a} = \CM(P_a,P_{a'}) = \KR(P_a,P_{a'})$ by \cref{cor:CM=KR}.
\end{proof}

\begin{proof}[Proof of \cref{prop:linear_QP}]

To illustrate the principle of the proof, we first address the simpler case where $\Phi = \Id$, and thus $p_0 = Q_{\ttx}$. Since $g_a$ is cyclically monotone by assumption, there exists a convex function $\varphi_a$ such that $g_a = \nabla \varphi_a$. Next, note that $\Jac(g_a) = \nabla^2 \varphi_a$, where $\varphi_a$ is convex, and that $\Jac(g_a) = (\Id - M^{\ttx})^{-1}$, according to \cref{prop:ctf_linear}. Therefore, $(\Id - M^{\ttx})^{-1}$ is symmetric positive semi-definite, and even positive definite as it is invertible. Consequently, there exists an invertible matrix $R$ such that $(\Id - M)^{-1} = R^T R$, which implies that $M^{\ttx} = \Id - R^{-1} (R^T)^{-1}$. Therefore, $M^{\ttx}$ is symmetric.

If $\Phi \neq \Id$ but is still diagonal, then one can apply the chain rule to obtain: 
$$\Jac(g_a)(\Phi(e)) \Jac{\Phi}(e) = \nabla^2 \varphi_a (e)$$ for every $e \in \dE$, where $\dE$ is the support of $p_0$. Next, note that $\Jac(g_a)(\Phi(e)) = (\Id - M^{\ttx})^{-1}$ and $\Jac{\Phi}(e) = \Diag(\phi'_1(e_1),\ldots,\phi'_d(e_d))$, where $\phi'_i(e_i) > 0$ for every $i \in [d]$. This leads to $$(\Id - M^{\ttx})^{-1} = \nabla^2 \varphi_a(e) \Diag(1/\phi'_1(e_1),\ldots,1/\phi'_d(e_d)).$$ Then, decomposing $\nabla^2 \varphi_a (e)$ as $R(e)^T R(e)$, where $R(e)$ is invertible, gives $$(\Id - M^{\ttx})^{-1} = R(e)^T R(e) \Diag(1/\phi'_1(e_1),\ldots,1/\phi'_d(e_d)),$$ and finally $M^{\ttx} = \Id - \Diag(\phi'_1(e_1),\ldots,\phi'_d(e_d)) R(e)^{-1} (R(e)^T)^{-1}$. Therefore, $M^{\ttx}$ is not necessarily symmetric, but satisfies $M^{\ttx}_{i,j} = 0 \iff M^{\ttx}_{j,i}=0$. 
\end{proof}

\begin{proof}[Proof of \cref{prop:QP_SCM}]
The proof follows the same principle as \cref{prop:linear_QP}, without relying on an explicit expression for $\Jac(g_a)$. This leads to $$\Jac{g_a}(\Phi(e)) = R(e)^T R(e) \Diag(1/\phi'_1(e_1),\ldots,1/\phi'_d(e_d)),$$ for every $e \in \dE$, which can be expressed as
\[
\Jac(g_a)(u_{\ttx}) = R(\Phi^{-1}(u_{\ttx}))^T R(\Phi^{-1}(u_{\ttx})) \Diag(1/\phi'_1(\Phi^{-1}(u_{\ttx})_1),\ldots,1/\phi'_d(\Phi^{-1}(u_{\ttx})_d))
\]
for every $u_{\ttx} \in \U_{\ttx}$. As such, $\Jac(g_a)(u_{\ttx})$ is not necessarily symmetric, but satisfies $\Jac(g_a)(u_{\ttx})_{i,j} = 0 \iff \Jac(g_a)(u_{\ttx})_{j,i}=0$. Therefore, by the assumption on $\Jac(g_a)$, $j \in \an(i) \iff i \in \an(j)$.
\end{proof}

\begin{proof}[Proof of \cref{prop:KR_ctf}]
Let $\M \in \SCM_{acy}$ and $a,a' \in \A$. According to \cref{thm:main_acy}, $C_{a' \leftarrow a} = \KR(P_a,P_{a'}) \iff g_a$ and $g_{a'}$ are diagonally comonotone. Furthermore, according to \cref{prop:acyclic}, $g_a(u_{\ttx})_i = f_i(a_{\pa(i)},\psi_i(a,u_{\an(i) \setminus \{i\}}),u_i)$ and $g_{a'}(u_{\ttx})_i = f_i(a'_{\pa(i)},\psi_i(a',u_{\an(i) \setminus \{i\}}),u_i)$. As such, if for every $x \in \X$ the functions $f_i(a_{\pa(i)},x_{\pa(i)},\cdot)$ and $f_i(a'_{\pa(i)},x_{\pa(i)},\cdot)$ are comonotone, then $C_{a' \leftarrow a} = \KR(P_a,P_{a'})$.
\end{proof}

\begin{proof}[Proof of \cref{thm:KR}]
Let $\M^\star := (\X \times \A, \G, \U^\star, Q^\star, f^\star)$ be an acyclic SCM with $(P_a)_{a \in \A}$ as interventional marginals. Since $\M^\star$ is acyclic, it is uniquely solvable with respect to any subset of variables. In particular, every solution $(X^\star,A^\star,U^\star)$ of $\M^\star$ share the same probability distribution $P^\star$.

The idea of the proof consists in modifying $\M^\star$ as to change the counterfactuals without altering the observational distribution and the graph. For every $i \in [d]$, define
\[
f_i : \A_{\pa(i)} \times \X_{\pa(i)} \times [0,1] \to \X_i, (a_{\pa(i)},x_{\pa(i)},u_i) \mapsto G_i(a_{\pa(i)},x_{\pa(i)},u_i),
\]
where $G_i(a_{\pa(i)},x_{\pa(i)},\cdot)$ is the quantile function of $\law{X^\star_i|A^\star_{\pa(i)}=a_{\pa(i)},X^\star_{\pa(i)}=x_{\pa(i)})}$ for every $(x,a) \in \X \times \A$, which is monotone increasing. Crucially, $G_i$ is jointly measurable, as a consequence of \citep[Theorem 2.1]{carlier2016vector}. Then, define $\M := (\X \times \A, \G, [0,1]^d \times \U^\star_{\tta}, \square \otimes Q^\star_{\tta}, f)$, where $f_{\tta} = f^\star_{\tta}$. Note that this construction can be seen as a monotone version of the generic noise-outsourcing procedure that we used in \cref{lem:markov}.

Next, we show that $\M$ meets the desired properties. To build a solution of $\M$, fix $(U_1,\ldots,U_d, U_{\tta})$ a vector of mutually independent random variables such that $(U_1,\ldots,U_d) \sim \square$ and $U_{\tta} \sim Q^\star_{\tta}$. Then, define $(X,A)$ recursively along the topological order induced by $\G$ as
\begin{align*}
    A &= f_{\tta}(X_{\pa(\tta)},U_{\tta}),\\
    X_i &= f_i(A_{\pa(i)},X_{\pa(i)},U_i) \text{ for every }i \in [d].
\end{align*}
Note that, by acyclicity, $U^\star_i \independent (A^\star_{\pa(i)},X^\star_{\pa(i)})$ and $U_i \independent (A_{\pa(i)},X_{\pa(i)})$ \citep[Remark 3.2]{cheridito2025optimal}. Therefore, the definition of $(G_i)^d_{i=1}$ leads to the following key identity: 
\begin{align*}
    \law{X_i | A_{\pa(i)} = a_{\pa(i)},X_{\pa(i)}=x_{\pa(i)}} &= \law{G_i(a_{\pa(i)},x_{\pa(i)},U_i)}\\
    &= \law{X^\star_i | A^\star_{\pa(i)} = a_{\pa(i)},X^\star_{\pa(i)}=x_{\pa(i)}},
\end{align*}
for every $i \in [d]$ and every $(x,a) \in \X \times \A$.

On this basis, we show that $\M$ admits $(P_a)_{a \in \A}$ as interventional marginals, that is, $\law{X_a} = P_a$ for every $a \in \A$. Let $a \in \A$ and compute using \cref{lem:markov},
\begin{align*}
    \rmd P_a(x_1,\ldots,x_d) :&= \rmd \bbP(X^\star_a=(x_1,\ldots,x_d))\\
    &= \prod^d_{i=1} \rmd \bbP(X^\star_{a,i} = x_i | A^\star_{\pa(i)} = a_{\pa(i)},X^\star_{\pa(i)}=x_{\pa(i)})\\
    &= \prod^d_{i=1} \rmd \bbP(X_{a,i} = x_i | A_{\pa(i)} = a_{\pa(i)},X_{\pa(i)}=x_{\pa(i)})\\
    &= \rmd \bbP(X_a=(x_1,\ldots,x_d)).
\end{align*}
To conclude, simply note that $\M$ meets the assumption of \cref{prop:KR_ctf}. 
\end{proof}

\end{document}